\documentclass[letterpaper, 10 pt, conference]{ieeeconf}  

\IEEEoverridecommandlockouts                              

\usepackage{censor}

\usepackage{graphics} 
\usepackage{epsfig} 
\usepackage{times} 
\usepackage{amsmath} 
\usepackage{amssymb}  

\usepackage{mathtools}          
\usepackage{mathrsfs}           
\usepackage{graphicx}           
\usepackage{subcaption}         
\usepackage[space]{grffile}     
\usepackage{url}                
\usepackage{lipsum}             
\usepackage{algorithm}
\usepackage[noend]{algpseudocode}
\usepackage{mdframed}
\usepackage{xcolor}
\usepackage{hyperref}
\makeatletter
\let\labelindent\@undefined
\makeatother
\usepackage{enumitem}

\title{\LARGE \bf
Training Fast Robot Policies with Slow Foundation Models
}

\author{Utsav Singh$^{1}$, Pramit Bhattacharyya$^{2}$, Vinay P. Namboodiri$^{3}$, Amrit Singh Bedi$^{1}$
\thanks{$^{1}$Utsav Singh and Amrit Singh Bedi are with the Department of Computer Science, University of Central Florida, Orlando, FL, USA
        {\tt\small utsav.singh@ucf.edu, amritbedi@ucf.edu}}%
\thanks{$^{2}$Pramit Bhattacharyya is with the Department of Computer Science and Engineering, IIT Kanpur, India
        {\tt\small pramitb@cse.iitk.ac.in}}%
\thanks{$^{3}$Vinay P. Namboodiri is with the Department of Computer Science, University of Bath, Bath, UK
        {\tt\small vpn22@bath.ac.uk}}%
}

\StopCensoring

\begin{document}
\bstctlcite{IEEEexample:BSTcontrol}

\maketitle
\thispagestyle{empty}
\pagestyle{empty}

\begin{abstract}

Continuous robotic control requires policies that execute with low latency and modest computational cost during deployment. Foundation models provide strong semantic and visual reasoning, but repeatedly querying a large model throughout deployment incurs substantial inference latency and compute requirements. Language-to-Reward (L2R) methods avoid this deployment-time cost by using large language models (LLMs) to synthesize rewards for training lightweight policies, but these rewards are generated without visually analyzing how the learned policy physically fails, and thus often lack physical grounding. We propose \emph{Visually-Grounded Reward Synthesis} (VGRS), which uses slow foundation models during training to produce fast robotic control policies. An LLM first synthesizes executable reward code from a natural-language instruction to train a lightweight hierarchical policy. When learning stalls, a frozen vision-language model (VLM) analyzes failed trajectories to provide failure mode diagnosis, which the LLM uses to rewrite and densify the reward. Since foundation models are used only during training, deployment requires only the learned policy. We perform experiments on simulated and real-world navigation and manipulation tasks, and show that VGRS achieves success rates above 55\% on challenging long-horizon tasks while deploying successfully to real robots.

\end{abstract}

\section{Introduction}
\label{sec:intro}
In practical continuous-control robotic tasks, the robot must
repeatedly observe its environment and generate the next
action within a short time window. The policy used for
deployment must therefore be fast, lightweight, and able to
run continuously on the computational resources available
on the robot. Compact control policies are well suited to
these requirements, but learning them for complex,
long-horizon tasks remains challenging. Successful behavior
often depends on intermediate geometric and temporal
conditions, such as approaching an object from the correct
direction, establishing a stable grasp, or completing one
subtask before moving to the next. These conditions are
difficult to capture using sparse task-success signals or
manually designed rewards.

\footnotetext[1]{%
  Accepted at IEEE/RSJ International Conference on Intelligent Robots and Systems (IROS) 2026, Pittsburgh, PA, USA. Corresponding author: {\href{mailto:utsav.singh@ucf.edu}{utsav.singh@ucf.edu}}%
}

Foundation models offer strong reasoning, planning, and code-generation abilities~\cite{huang2022language,ahn2022can,
liang2023code}, and are a natural source of the semantic and visual knowledge required by robotic control tasks. One way to use them is to deploy a vision-language-action (VLA) model~\cite{zitkovich2023rt,kimopenvla} that maps observations and instructions directly to actions. However, this requires
the large model to remain part of the deployed controller,
introducing repeated inference cost throughout execution.
Thus, although foundation models provide rich visual reasoning, their deployment profile is considerably heavier than that of a
compact policy designed specifically for continuous control.

\begin{figure}[t]
\centering
\includegraphics[width=0.8\columnwidth]{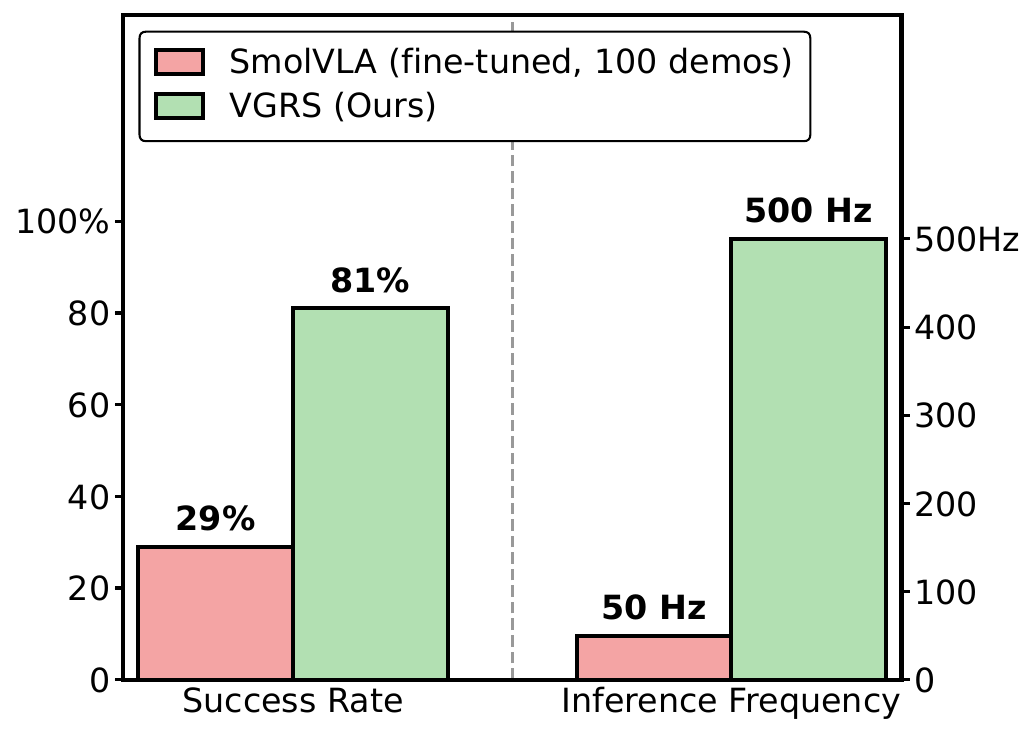}
\caption{\textbf{Fast policies with slow foundation models.}
On the Pick and Place task, SmolVLA fine-tuned on 100 expert demonstrations achieves success rate of 29\%, but the inference frequency remains at 50~Hz as the VLA backbone must still be queried at every inference step. In contrast, VGRS restricts foundation models to training and deploys a lightweight policy, achieving 81\% success at approx 500~Hz.}

\label{fig:vgrs_comparison}
\end{figure}

A natural alternative is to use foundation-model reasoning
while the policy is being trained, rather than every time the
robot must act. Prior \emph{Language-to-Reward} methods
follow this approach by using an LLM to translate a
natural-language task description into an executable reward
function~\cite{yu2023language,ma2023eureka}. Reinforcement
learning or trajectory optimization can then optimize this
reward to produce a lightweight policy. Once training is
complete, the policy can control the robot without querying
the LLM during deployment.

However, in existing L2R based methods, the reward is not grounded in the physical execution of the policy. Concretely, these methods can produce a
reward that is semantically consistent with the task
description without visually examining how the
learned policy behaves in the environment. As a result, the
reward may specify the desired outcome while overlooking
the physical conditions needed to achieve it. For example, a
reward may encourage the robot to lift an object without
ensuring that the gripper has first established a stable
grasp. The policy may then repeatedly collide with the
object or begin lifting too early, even though it is
optimizing a reward that appears reasonable from the task
description alone.

This reveals a limitation in the current training pipeline.
Direct VLA control retains visual reasoning during execution
but also retains repeated large-model inference at
deployment. L2R produces a lightweight deployable policy,
but its reward-refinement process is not informed by visual
evidence of why that policy fails. A desirable approach
would combine the strengths of both. It would use visual and
semantic reasoning to improve the reward during training,
while leaving only a lightweight policy to control the robot
at deployment.

\begin{figure*}[t]
\centering
\includegraphics[width=\textwidth]{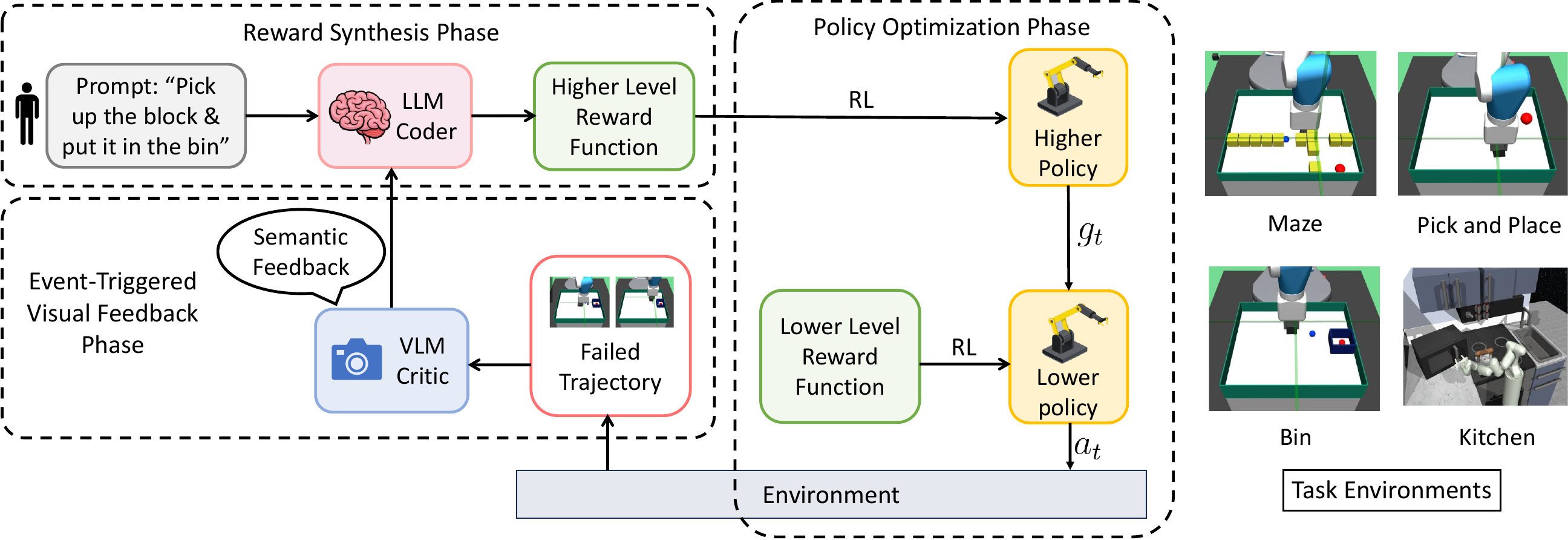}
\caption{\textbf{Overview of Visually-Grounded Reward Synthesis (VGRS) framework.} An LLM first converts a natural-language instruction into executable reward code, which is used to train a lightweight hierarchical control policy. When learning stagnates, rollout frames from failed trajectories are passed to a frozen VLM, which identifies the likely failure mode and provides semantic feedback. Conditioned on this feedback, the LLM rewrites and densifies the reward code, yielding a more grounded supervisory signal for policy learning. At deployment, only the lightweight policy is used for control.}
\label{fig:explain_method}
\end{figure*}

To this end, we propose \emph{Visually-Grounded Reward Synthesis} (VGRS), which
closes the loop between policy execution and reward design.
VGRS first uses an LLM to generate executable reward code
from a natural-language instruction. The reward is used to
train a lightweight hierarchical policy for the long-horizon
task. When policy improvement stagnates, a frozen VLM
examines frames from failed rollouts and identifies the likely
physical cause of failure. The resulting visual diagnosis is
then provided to the LLM, which revises the
reward code before policy learning continues. Figure~\ref{fig:vgrs_comparison} provides a direct comparison against a VLA baseline, and Figure~\ref{fig:explain_method} provides an overview of the VGRS approach.
%
%
%
%
\noindent Our main contributions are as follows:
\begin{itemize}[leftmargin=1em, labelsep=0.3em, itemindent=0pt]

\item We introduce \textbf{Visually-Grounded Reward
Synthesis (VGRS)}, a framework that closes the loop between policy execution and reward design by grounding executable reward code in visual evidence of how the learned policy physically fails, yielding fast, lightweight hierarchical policies that query no foundation model at deployment.

\item We propose an \textbf{event-triggered visual reward
refinement} mechanism in which a frozen VLM diagnoses
physical failure modes from rollout frames of failed
trajectories, and the resulting feedback conditions an LLM
to iteratively rewrite and densify the executable reward
code.

\item We demonstrate VGRS on simulated and real-world
navigation and manipulation tasks, showing success rates
above 55\% on challenging long-horizon tasks, and successful
transfer to a physical robot arm.

\end{itemize}

\section{RELATED WORK}
\label{sec:related}

\noindent \textbf{Language-to-actions.}
Language-conditioned control has evolved from early symbolic parsing~\cite{kress2008translating, matuszek2013learning} to end-to-end policies trained via behavioral cloning~\cite{brohan2023rt}, offline RL~\cite{ebert2021bridge}, or goal-conditioned RL~\cite{jiang2019language}. Vision-Language-Action (VLA) models~\cite{kimopenvla, brohan2023rt} extend this paradigm by conditioning directly on pixel observations and language instructions to predict primitive actions, achieving strong generalization across manipulation tasks. However, VLAs require extensive robot-specific fine-tuning on thousands of expert trajectories~\cite{kimopenvla} and incur prohibitive inference latency for real-time control~\cite{brohan2023rt}. Our approach introduces visual grounding into reward synthesis without querying a foundation model at deployment, by using VLMs during training to generate failure feedback for reward synthesis, thus removing the need to fine-tune a large foundation model.

\noindent \textbf{Language-to-code.}
Prior works like Code-as-Policies (CaP)~\cite{liang2023code} and ProgPrompt~\cite{singh2023progprompt} exploit LLMs' code-generation capabilities to synthesize executable robot programs from natural language, enabling hierarchical composition over primitive libraries. These approaches excel at structured reasoning but require manual primitive libraries to bridge language to low-level actions~\cite{liang2023code}, limiting scalability to novel tasks or embodiments. Our approach eliminates this dependency by using LLMs to generate reward functions rather than policy code, enabling downstream RL optimization without predefined skills.

\noindent \textbf{Language-to-rewards.} Language-to-Reward (L2R) frameworks translate instructions into reward signals for RL optimization~\cite{yu2023language, ma2023eureka, kwon2023reward, sharma2022correcting}. Eureka~\cite{ma2023eureka} uses LLMs to evolve reward code through iterative refinement, while RewardCoder~\cite{yu2023language} decomposes instructions into structured motion descriptors before reward parameterization. These methods construct and refine rewards from task text, environment code, and scalar training statistics. Auto MC-Reward~\cite{automcreward} additionally uses an LLM trajectory analyzer to summarize failure causes from collected trajectories and revise executable reward code. However, its analyzer operates on symbolic game state and statistics rather than visual observations, so the reward remains visually unaware of physical failure modes. VGRS addresses this by introducing an event-triggered VLM visual critic that diagnoses physical failure modes directly from rendered rollout frames in continuous-control robotics, capturing geometric and temporal constraints that are not represented in symbolic state. The resulting feedback conditions the LLM reward synthesizer to correct the reward during training, and at deployment the learned policy queries no foundation model.

\noindent \textbf{Hierarchical reinforcement learning.}
Hierarchical Reinforcement Learning (HRL) is a promising approach that decomposes long-horizon tasks into multi-level temporal abstractions to mitigate sparse reward exploration and long-term credit assignment challenges~\cite{barto2003recent, nachum2018data, levy2019learning}. Initial work formalized HRL through options~\cite{sutton1999between}, while recent advances have focused on automated skill discovery~\cite{eysenbach2018diversity} and off-policy stability~\cite{nachum2018data, levy2019learning}. Despite these advances, HRL suffers from training instability: as the low-level policy improves, the high-level reward distribution becomes non-stationary, destabilizing training~\cite{nachum2018data, levy2019learning}. Prior solutions include hindsight relabeling~\cite{nachum2018data}, leveraging preference-based learning~\cite{singh2024piper,singhdirect} or expert demonstrations~\cite{lynch2019play,singh2023pear,singh2023crisp}. 
In VGRS, the synthesized reward is decoupled from the current
performance of the lower-level policy, which reduces instability in hierarchical learning, while enabling adaptation to novel tasks through language.
\section{METHODOLOGY}
\label{sec:method}
\subsection{Preliminaries}
Our framework converts natural-language user instructions into robotic behaviors by utilizing executable reward functions as the interface between high-level semantic intent and low-level continuous control. By synthesizing a task-conditioned reward function from language to train the control policy, we ground semantic objectives in physically executable actions.

We define the control task as a Markov Decision Process (MDP), characterized by the tuple $\mathcal{M} = (\mathcal{S}, \mathcal{A}, P, R, \gamma, p_0)$, where $\mathcal{S}$ is the continuous state space, $\mathcal{A}$ is the continuous action space, $P(s_{t+1} \mid s_t, a_t)$ represents the transition dynamics, $R: \mathcal{S} \times \mathcal{A} \times \mathcal{G} \to \mathbb{R}$ is the reward function, $\gamma \in [0,1)$ is the discount factor, and $p_0$ is the initial state distribution.

To tackle long-horizon tasks, we adopt a two-level HRL architecture. At the higher level, a subgoal policy $\pi_H(g_t \mid s_t)$ proposes subgoals $g_t \in \mathcal{G}$ every $k$ environment steps, conditioned on the current state $s_t$. The objective of this high-level policy is to maximize the expected task return $J_H = \mathbb{E} \left[ \sum_{c=0}^\infty \gamma_H^c R_H(s_{c \cdot k}, g_{c \cdot k}) \right]$, where $R_H$ is the semantically grounded reward synthesized by the LLM. At the lower level, a primitive policy $\pi_L(a_t \mid s_t, g_t)$ executes continuous actions at every environment step to achieve the subgoal $g_t$. This low-level policy maximizes its own intrinsic tracking reward $J_L = \mathbb{E} \left[ \sum_t \gamma_L^t R^L_t \right]$, where $R^L_t$ strictly penalizes the distance to the active subgoal (e.g., $R^L_t = -\mathbf{1}_{\|s_{t+1} - g_t\| > \epsilon}$).

Rather than manually designing the high-level reward function $R_H$, which evaluates the quality and physical feasibility of the proposed subgoal $g_t$, we formulate it as an executable code block synthesized by an LLM. We assume this reward computes a sum of continuous distances and heuristic penalties:
\begin{align}
\label{eqn:eqn_1}
R_H(s_t, g_t) = -\sum_{i=0}^{M} w_i \cdot \| \rho_i(s_t, g_t; \psi_i) \|_2
\end{align}
where $w_i \in \mathbb{R}^+$ is a non-negative weight, $\rho_i$ is a residual distance or heuristic term that achieves optimality when $\rho_i = 0$, and $\psi_i$ represents the parameters of the $i$-th residual term (e.g., spatial coordinates, gripper states).

In principle, one could manually design task-specific residual terms $\rho_i$ to train $\pi_H$. For example, a residual term $\rho_{\text{dist}} = g_t - \psi_{\text{target}}$ could penalize the distance between a subgoal and a target object. However, manually engineering these residuals requires extensive domain expertise, scales poorly to novel tasks, and fails to dynamically adapt to physical execution failures. Instead, we use a set of generic environment APIs (e.g., \texttt{get\_goal\_pos()} which gives the goal position, and \texttt{goal\_distance(current\_position)}, which gives the distance between current position and the goal, etc), and leverage the coding capabilities of LLMs, guided by the visual grounding of VLMs to autonomously compose, parameterize, and update these residual terms as an executable reward script. We provide the full LLM reward synthesizer and VLM Critic prompt templates in Appendix~\ref{appendix:templates}.

\subsection{Visually-Grounded Reward Synthesis (VGRS)}
Our objective is to learn a lightweight hierarchical control policy that solves long-horizon tasks specified by language instructions and executes at high frequency without querying any foundation model at deployment. To achieve this, VGRS uses LLMs and Vision-Language Models (VLMs) strictly as training-time reward synthesizers, yielding visually grounded reward functions that iteratively adapt to physical task constraints.

As illustrated in Figure~\ref{fig:explain_method}, our framework operates in two core phases:
$(i)$ \textit{Initial Reward Synthesis}: An LLM translates a semantic task instruction into an initial executable reward function using generic environment APIs.
$(ii)$ \textit{Event-Triggered Visual Feedback:} If the policy's success rate stagnates during training, a frozen VLM acts as a visual critic. It analyzes trajectory rollout frames to diagnose physical execution failures and generates natural-language feedback. This feedback is provided as context to the LLM, enabling it to rewrite and densify the reward function.

Throughout this process, the high-level policy is continuously trained via reinforcement learning to maximize this iteratively refined reward, learning to propose effective subgoals for the low-level primitive. This enables the learned policy to execute complex, visually grounded behaviors at high control frequencies without inference bottlenecks. We now discuss these phases in detail.

\subsection{Language-Guided Initial Reward Synthesis Phase}
We leverage a pre-trained LLM, $\Phi_{\text{LLM}}$, to act as a zero-shot reward coder. Given the natural language task instruction $\mathcal{T}$ and a system prompt $\mathcal{P}_{\text{env}}$ explaining the generic API structure, the LLM synthesizes an initial, parameterized Python reward program $R_H^{(0)}$:
\begin{align}
\label{eqn:eqn_2}
    R_H^{(0)} \leftarrow \Phi_{\text{LLM}}(\mathcal{T}, \mathcal{P}_{\text{env}})
\end{align}
This acts as a deterministic mapping $(s_t, g_t) \rightarrow \mathbb{R}$, computing a dense reward signal for any proposed subgoal based on the residual structure defined in Equation~\ref{eqn:eqn_1}. Notably, because this reward is generated by the LLM, the higher-level reward function is decoupled from the current performance of the low-level primitive $\pi_L$, which stabilizes HRL training.

\subsection{Event-Triggered Visual Feedback Phase}
Although LLMs can synthesize syntactically correct reward code, they lack direct access to visual evidence of policy execution and therefore cannot reliably infer missing physical constraints. Consequently, the initially synthesized reward $R_H^{(0)}$ often suffers from a \emph{grounding gap}, due to which it may fail to penalize physically infeasible subgoals or overlook specific hardware bottlenecks that only emerge during task execution. In order to introduce environment-specific physical grounding into the reward, we introduce an event-triggered visual feedback generation phase. In this phase, we monitor the moving average $\eta_w$ of the policy's success rate on past training episodes. We define a stagnation trigger $\alpha$:
\begin{align}
    \alpha = \mathbf{1}[\nabla \eta_w \approx 0 \text{ and } \eta_w < \eta_{\text{threshold}}]
\end{align}
When $\alpha = 1$, we extract a sequence of visual frames, $I_{\text{fail}}$, from the most recent failed trajectory and pass them to a frozen VLM, $\Psi_{\text{VLM}}$. The VLM acts as a visual critic, prompted to analyze the physical cause of failure relative to the instruction $\mathcal{T}$, which outputs feedback in the form of a semantic description $\mathcal{F}$ clearly explaining the failure mode and its causes (e.g., \textit{``The robot successfully reached the object, but the gripper closed prematurely before grasping the object''}):
\begin{align}
    \mathcal{F} = \Psi_{\text{VLM}}(I_{\text{fail}}, \mathcal{T})
\end{align}
This textual feedback compresses complex, high-dimensional visual failure modes into a compact semantic diagnosis for the LLM. The LLM is then conditioned on this feedback, the original task $\mathcal{T}$, the previous reward logic $R_H^{(m)}$, and the new visual feedback $\mathcal{F}$. Based on this context, the LLM rewrites the reward code to synthesize an updated, visually grounded dense reward function $R_H^{(m+1)}$:
\begin{align}
    R_H^{(m+1)} \leftarrow \Phi_{\text{LLM}}(\mathcal{T}, P_{\mathrm{env}}, \mathcal{F}, R_H^{(m)})
\end{align}
This iterative process allows the reward function to dynamically adapt to the agent's physical limitations and the constraints of the environment.

\subsection{Policy Optimization}
The hierarchical policies are optimized concurrently via off-policy hierarchical reinforcement learning:

\textit{Lower-Level Optimization.} The primitive policy $\pi_L$ is trained to reach the subgoals $g_t$ generated by $\pi_H$, supervised strictly by the intrinsic distance penalty $R_L(s_{t+1}, g_t) = -\mathbf{1}\{\|s_{t+1}-g_t\|>\epsilon\}$.

\textit{Higher-Level Optimization.} The subgoal policy $\pi_H$ maximizes the expected return defined by the latest LLM-generated reward function. To maintain off-policy learning stability, past high-level transitions in the replay buffer are dynamically relabeled using the current reward function $R_H^{(m)}(s_{t+k}, g_t)$.

Crucially, our framework isolates the computational burden of foundation models entirely to the training phase. At inference time, neither the VLM nor the LLM is queried. This avoids the latency bottlenecks typical of foundation models, enabling the lightweight hierarchical policy to execute complex, visually grounded tasks at high control frequencies. The full pseudo-code is summarised in Appendix~\ref{appendix:algorithm} Algorithm~\ref{alg:vgrs}.

\section{EXPERIMENTS}
\label{sec:experiment}

\begin{figure*}[t]
\centering
\subfloat[][Maze navigation]{\includegraphics[scale=0.26]{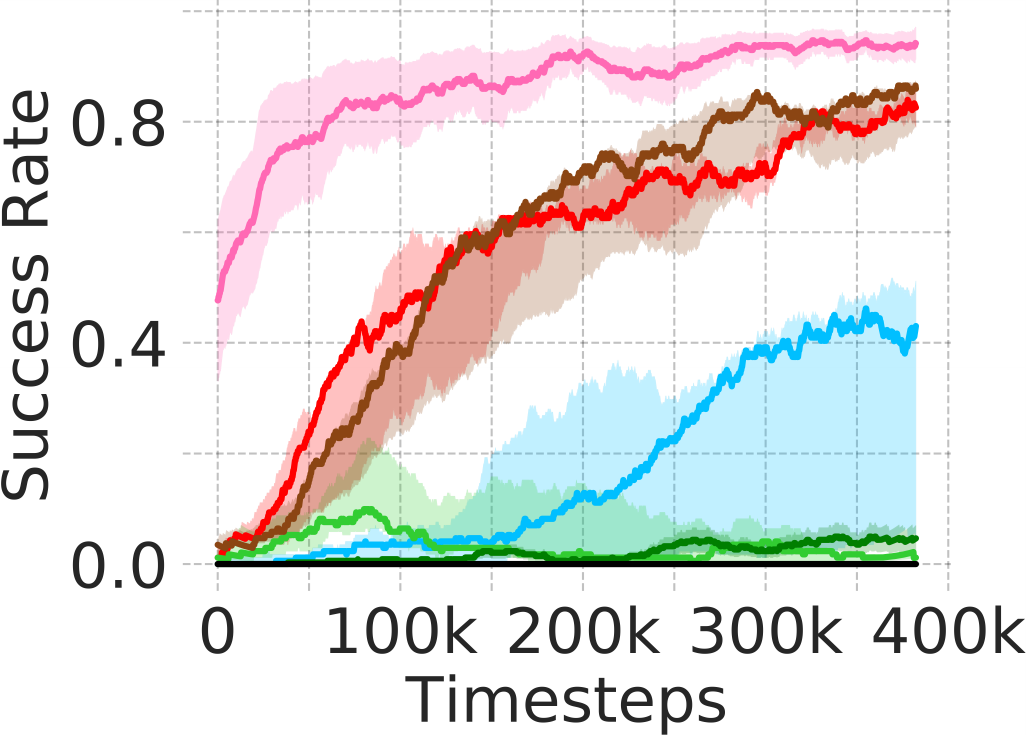}}
\subfloat[][Pick and place]{\includegraphics[scale=0.26]{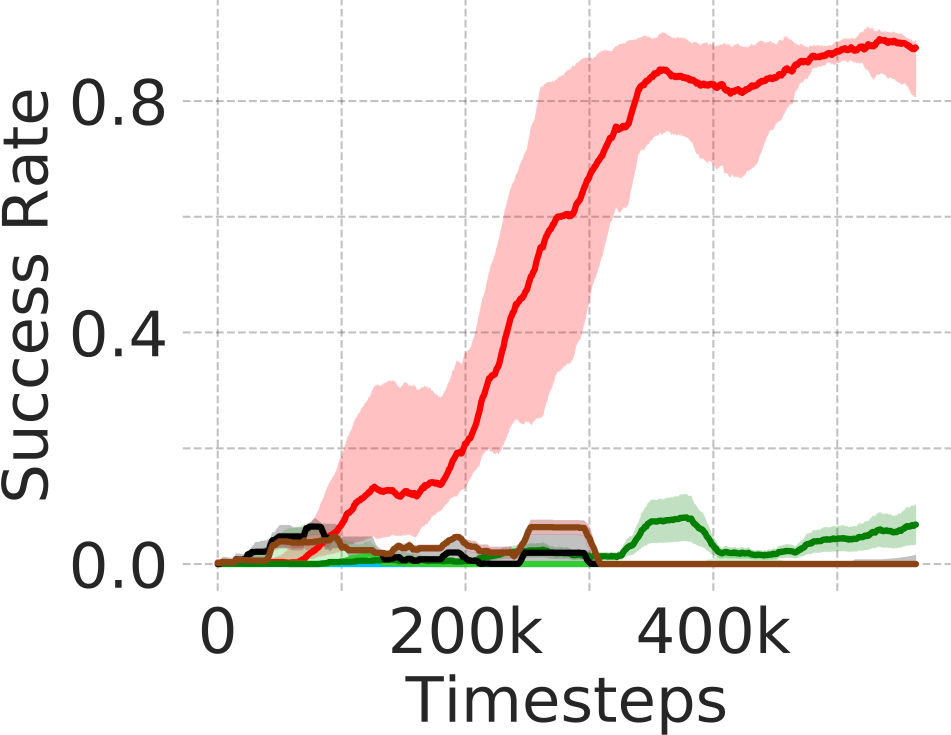}}
\subfloat[][Bin]{\includegraphics[scale=0.26]{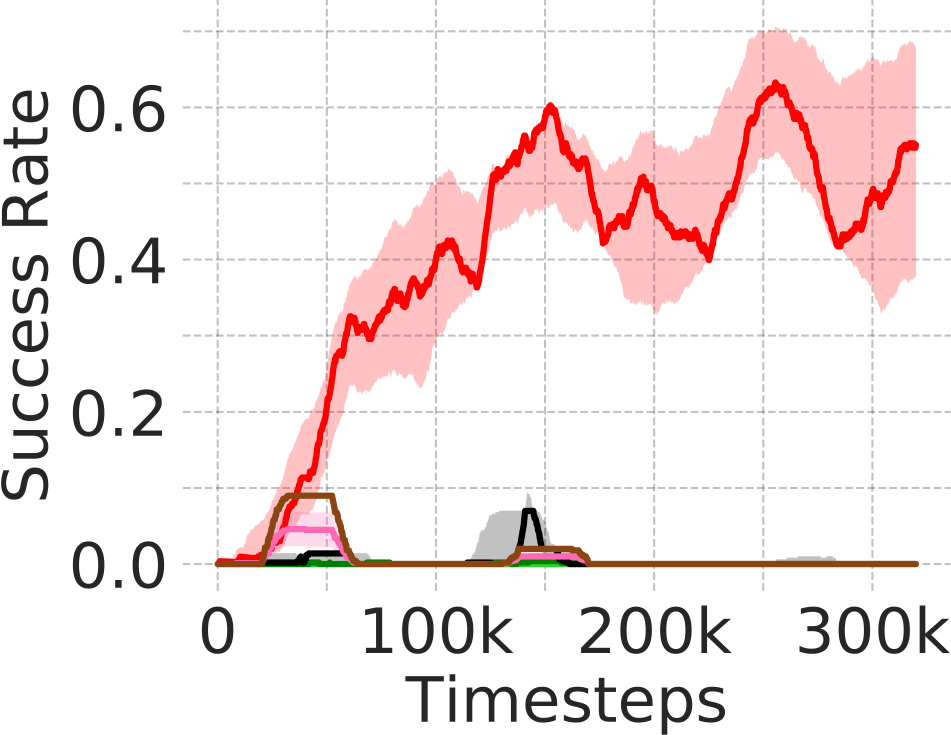}}
\subfloat[][Kitchen]{\includegraphics[scale=0.26]{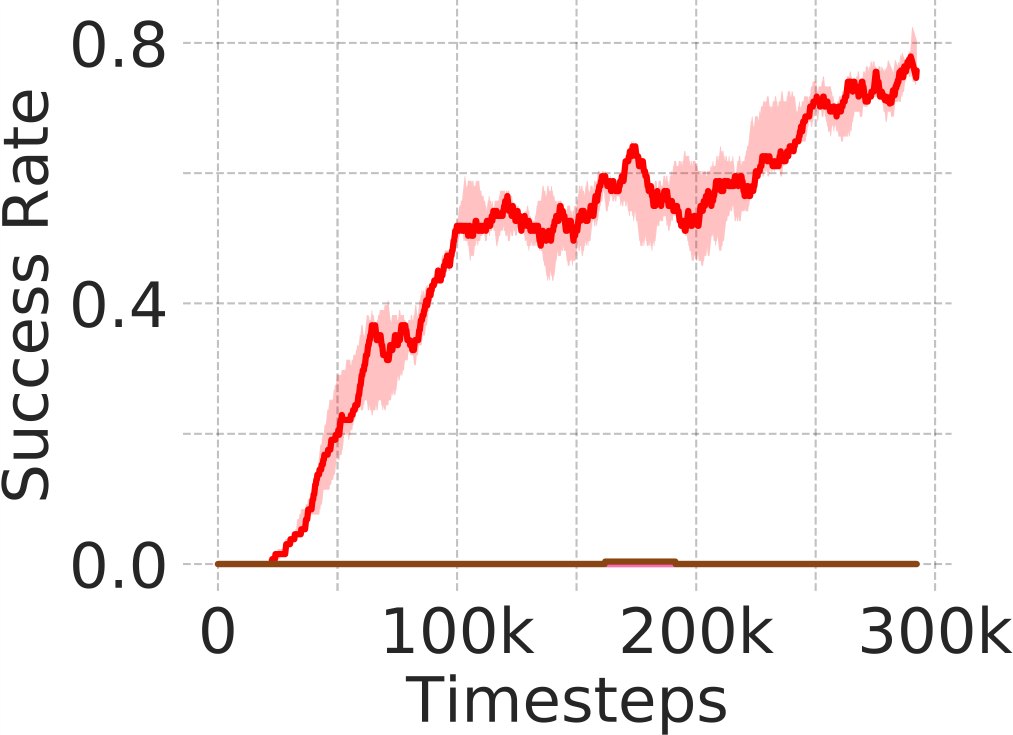}}
\\
{\includegraphics[scale=0.4]{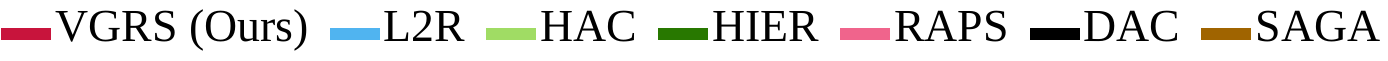}}
\caption{\textbf{Success Rate Comparison.} In this figure, we present the success rates over the course of training for four sparse-reward environments: (a) maze navigation, (b) pick-and-Place, (c) bin, and (d) franka Kitchen. The solid lines and shaded regions represent the mean and standard deviation across 5 random seeds. Our proposed VGRS framework is competitive with strong baselines on maze navigation and substantially outperforms them on the three more challenging manipulation tasks, across both single-level foundation model approaches and standard hierarchical baselines. \textit{(a)} While single-level language baselines (L2R) and traditional HRL methods (SAGA, RAPS) perform well on the simpler Maze task, their performance deteriorates sharply on harder manipulation tasks, whereas our approach is able to maintain high success rates in harder tasks. \textit{(b, c)} On pick-and-place and bin, standard HRL baselines like HAC and HIER fail to make progress in sparse reward scenarios, whereas VGRS leverages VLM-densified rewards to stabilize and accelerate hierarchical learning. \textit{(d)} In the most complex, multi-stage franka kitchen task, methods relying on fixed action primitives (RAPS) or static, non-updating LLM rewards (L2R) fail completely. In contrast, VGRS uses VLM-refined, visually grounded rewards, achieving superior final performance and sample efficiency.}
\label{fig:success_rate_comparison}
\end{figure*}

In our experiments, we aim to answer the following research questions:
\begin{itemize}[leftmargin=1em, labelsep=0.3em, itemindent=0pt]
    \item \textbf{RQ1:} How does VGRS compare against prior language-guided, hierarchical, and single-level baselines in terms of task success rate and inference latency?
    \item \textbf{RQ2:} Does the event-triggered visual feedback loop effectively densify the reward signal compared to standard hierarchical RL?
    \item \textbf{RQ3:} Is a decoupled VLM-to-LLM pipeline superior to using a single, unified VLM for code generation?
    \item \textbf{RQ4:} How critical is the visual critique mechanism for overcoming physical failure modes? 
\item \textbf{RQ5:} Why not use foundation models directly for control in our target environment?
    \item \textbf{RQ6:} Can policies trained via VGRS transfer to real-world robotic setups?
\end{itemize}
\subsection{Implementation Details} 
We evaluate VGRS across a suite of continuous robotic navigation and manipulation tasks, including maze navigation, pick-and-place~\cite{levy2019learning}, bin, and franka kitchen~\cite{lynch2019play}. These environments are characterized by complex, long-horizon objectives and sparse rewards, where the agent only receives a success signal upon full task completion (e.g., reaching within $\delta$ radius of a target). In the maze task, intermediate waypoints serve as subgoals; for more complex tasks like franka kitchen, full intermediate states are used as subgoals. The high-level and low-level policies are modeled as three-layer multi-layer perceptrons (MLPs) with 512 neurons per layer. Both levels are optimized using Soft Actor-Critic (SAC)~\cite{haarnoja2018soft} with the Adam optimizer. For the foundation models, we utilize llama-3.3-70b-versatile~\cite{grattafiori2024llama} as the LLM ($\Phi_{\text{LLM}}$) and llama-4-scout-17b-16e-instruct~\cite{meta2025llama} as the VLM ($\Psi_{\text{VLM}}$). We carefully tune hyperparameters via grid search across all baselines to ensure fair comparison. For challenging tasks such as pick and place and kitchen, we incorporate a single demonstration and an imitation learning objective at the lower level; no demonstrations are used for maze navigation to maintain consistent evaluation. In the Appendix, we procide the implementation and environment details (Appendix~\ref{sec:implementation_details} and~\ref{sec:environment_details}), additional hyperparameters (Appendix Table~\ref{tab:hierarchical_hyperparams}), and qualitative visualizations (Appendix ~\ref{appendix:qualitative_vizualizations}.

\subsection{RQ1: Success Rate Performance and Inference Latency Comparison}

To comprehensively evaluate VGRS, we compare it against a diverse set of baselines spanning language-guided reward generation, HRL, and imitation learning in Figure~\ref{fig:success_rate_comparison}, where the results are averaged over 5 seeds.

\noindent \textbf{L2R.} The L2R approach~\cite{yu2023language} translates natural language 
instructions into reward parameters via a reward translator, and originally uses an MPC 
controller~\cite{howell2022predictive}. For a fair comparison, our baseline adopts the same 
translator but replaces MPC with a Soft Actor-Critic (SAC)~\cite{haarnoja2018soft} 
agent. Beyond lacking a hierarchical structure for temporal abstraction, L2R synthesizes its 
reward statically prior to training and never updates it based on physical execution failures. 
This single-level, static setup performs well on simpler maze tasks, possibly because for such simple tasks, statically generated rewards are enough to solve the task. However, L2R struggles 
significantly in harder sparse-reward domains like pick and place, bin and franka kitchen, where VGRS is able to consistently outperform L2R, by combining hierarchical subgoal decomposition with dynamic, visually grounded reward refinement.

\noindent \textbf{SAGA.} SAGA~\cite{wang2023state} is a hierarchical method that uses a 
state-conditioned discriminator to align subgoals with the low-level policy. While effective 
on maze tasks, SAGA performance deteriorates on harder tasks due to unstable hierarchical training. In contrast, VGRS outperforms this baseline by leveraging LLM-guided reward relabeling, which provides a stationary, semantically meaningful reward signal decoupled from the evolving low-level policy, thus stabilizing hierarchical training.

\noindent \textbf{RAPS.} The RAPS baseline~\cite{DBLP:journals/corr/abs-2110-15360} uses a 
library of predefined robot action primitives controlled by the higher-level policy. While 
RAPS performs adequately on simpler maze tasks, it also struggles on harder tasks, 
suggesting that methods relying on fixed primitives lack the flexibility required for complex 
objectives. In contrast, VGRS autonomously discovers effective dynamic reward functions to optimize the subgoal policy and achieves strong results on such tasks.

\begin{figure*}[t]
\centering

\subfloat[][Maze navigation]{\includegraphics[scale=0.25]{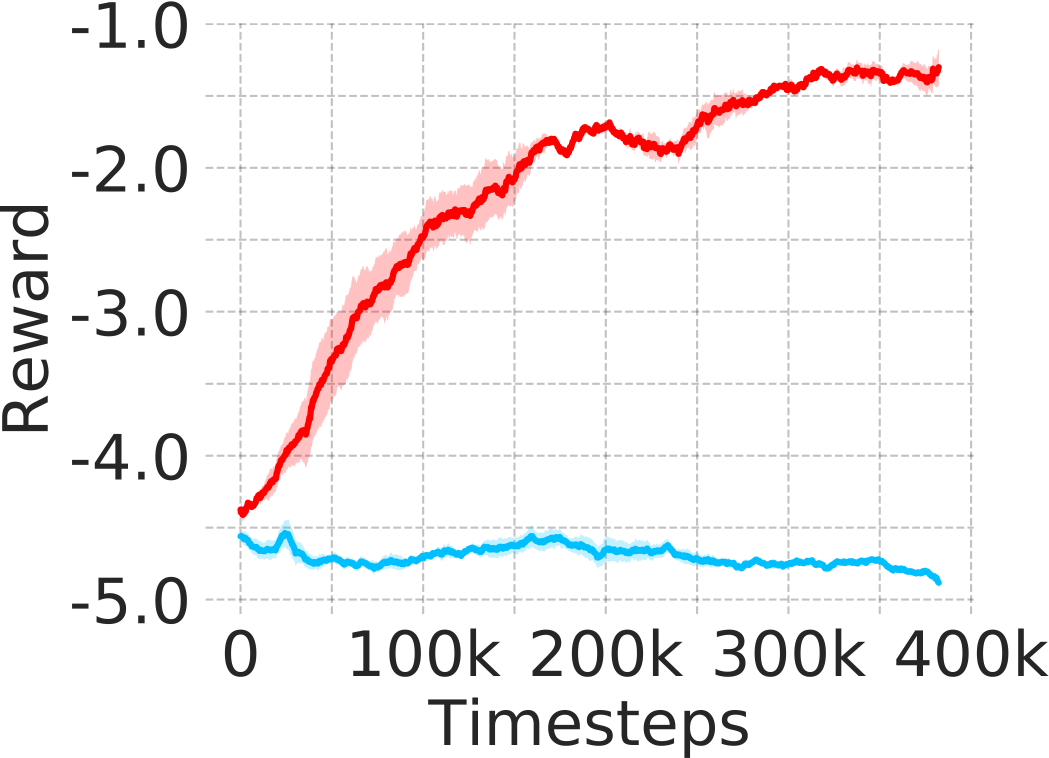}}
\subfloat[][Pick and place]{\includegraphics[scale=0.25]{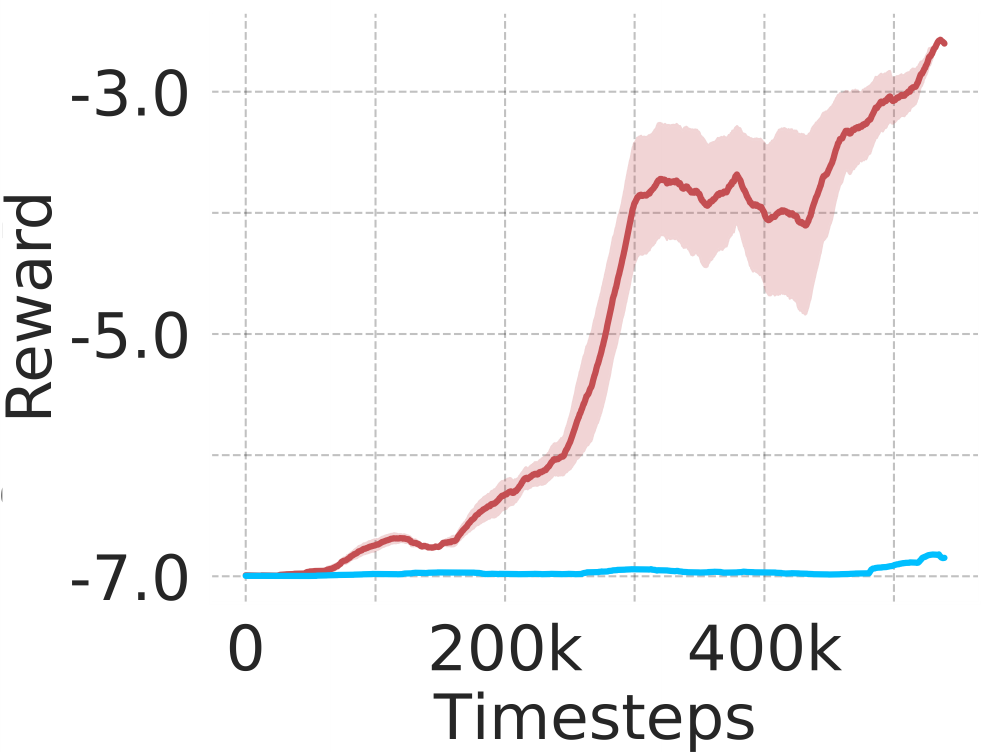}}
\subfloat[][Bin]{\includegraphics[scale=0.25]{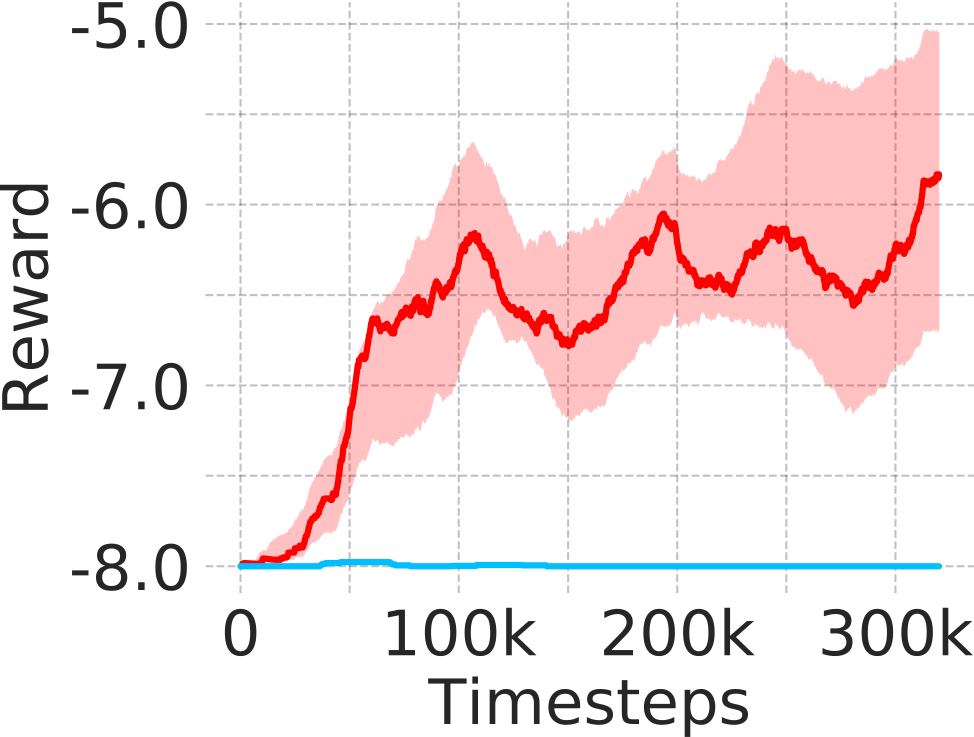}}
\subfloat[][Kitchen]{\includegraphics[scale=0.25]{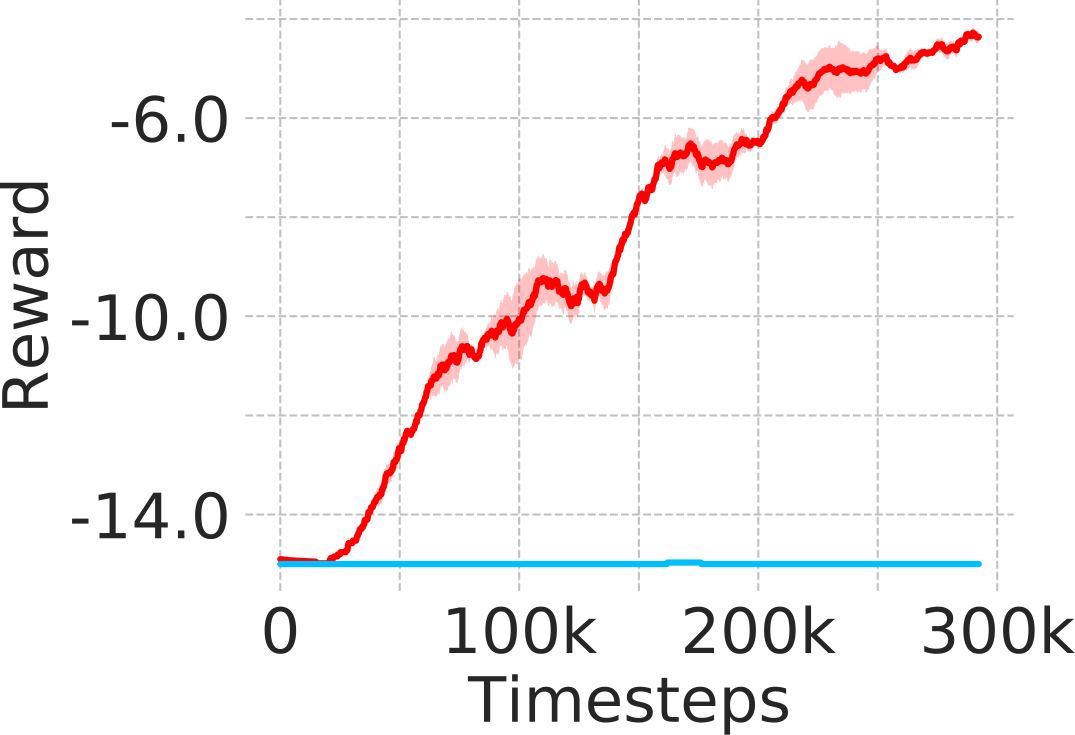}}
\\
{\includegraphics[scale=0.33]{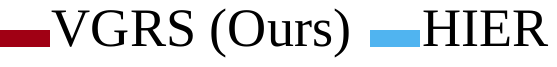}}
\caption{\textbf{Reward Densification Analysis.} This figure shows the average higher level rewards during training. The solid line and shaded regions represent the mean and standard deviation across 5 seeds. Compared to the sparse rewards of the vanilla HIER baseline, VGRS generates a significantly denser reward signal. This demonstrates the effectiveness of the VLM visual critic in diagnosing physical failure modes and guiding the LLM to dynamically densify the high-level rewards.}
\label{fig:reward_dense}
\end{figure*}

\noindent \textbf{HAC.} HAC~\cite{levy2019learning} attempts to stabilize HRL by 
assuming an optimal lower-level policy, and relabels the higher-level replay buffer using 
subgoal relabeling. Despite this, our experiments show VGRS consistently surpasses HAC. We attribute this to the fact that VGRS's LLM-generated reward relabeling produces stable, goal-aligned rewards for the higher-level policy without requiring the unrealistic assumption of a pre-trained optimal lower-level primitive, providing a more practical and robust mechanism for stabilizing HRL. Note that HIRO~\cite{nachum2018data} is another method that stabilizes HRL; since HAC has been empirically shown to outperform HIRO, we use HAC as the representative baseline.

\noindent \textbf{HIER.} The HIER baseline is a standard hierarchical SAC model where the higher-level reward is the cumulative sum of environment rewards across $k$ steps. VGRS outperforms HIER across all tasks, underscoring the critical role of  language-guided reward generation in stabilizing HRL.

\noindent \textbf{DAC.} Finally, we evaluate a Discriminator-Actor-Critic (DAC)~\cite{dac1} baseline provided with the same single 
human demonstration used to bootstrap our lower-level primitive. DAC fails to show any 
significant progress, demonstrating that VGRS outperforms single-level imitation baselines even when they have access to privileged expert demonstration data.

\noindent \textbf{Inference latency comparison.} 
By restricting foundation model queries entirely to the training phase, we avoid the prohibitive latency bottlenecks associated with deploying large models in robotic tasks. Approaches like Code as Policies~\cite{liang2023code} are fundamentally bottlenecked by token generation speeds (typically around 1 Hz with \texttt{llama-3.3-70b-versatile} model). Similarly, continuous control methods that deploy off-the-shelf VLAs require computationally heavy forward passes at every step, generally limiting inference frequencies (typically around 20-50 Hz). In contrast, since VGRS distills semantic reasoning and visual grounding into a lightweight multi-layer perceptron (MLP) hierarchy during training, the network's forward pass at deployment is computationally trivial (typically around 500 Hz for two level hierarchy, with three MLP layers consisting of 512 neurons each), enabling highly reactive, closed-loop continuous control.

\subsection{RQ2: Reward Densification Analysis}
To verify that our VLM-guided reward code generation process actually densifies the reward, we plot the average higher-level policy rewards over the course of training (Figure~\ref{fig:reward_dense}). Compared to the vanilla HIER baseline which relies strictly on sparse $k$-step environmental rewards, VGRS produces a significantly denser and structurally richer reward signal. The plots demonstrate that the VLM-based visual critic identifies the failure modes and provides feedback to the LLM, which in turn generates a dense reward function for the higher-level policy.

\subsection{RQ3: Comparing our VLM to LLM pipeline vs. Single VLM Baseline for Reward Code Generation} 
\label{sec:rq3}
A natural alternative to our decoupled approach that uses a VLM for failure mode diagnosis and an LLM for reward code generation, is to use a single VLM to both diagnose the failure and write the updated reward code. While Vision-Language Models (VLMs) demonstrate strong semantic visual reasoning on standard visual question-answering benchmarks, their structural and logical code-generation capabilities remain inferior to those of specialized Large Language Models (LLMs) \cite{jiang2025viscodex}. We found that when using a single VLM to write code directly based on the failure trajectory images, the single VLM baseline frequently generates syntactically incorrect APIs, syntax errors or mathematically unstable reward coefficients. For instance, in the pick and place task, the VLM-only baseline generated code that hallucinated non-existent blocks in the environment, while also demonstrating mathematical and dimensionality errors. By contrast, our decoupled approach uses the VLM output as a semantic critique generator ($\Psi_{\text{VLM}}$), allowing a dedicated, code-focused LLM ($\Phi_{\text{LLM}}$) to maintain strict adherence to spatial coordinates, API constraints, and syntax, yielding reliable and mathematically correct reward functions. In Appendix~\ref{appendix:vlm_only}, we provide an example of reward code generation comparing our VLM and LLM pipeline versus a VLM only baseline, clearly showing that the VLM only baseline generates incorrect code compared to our approach.

\begin{figure}[t]
\centering
\subfloat[][Static Reward]{\includegraphics[scale=0.2]{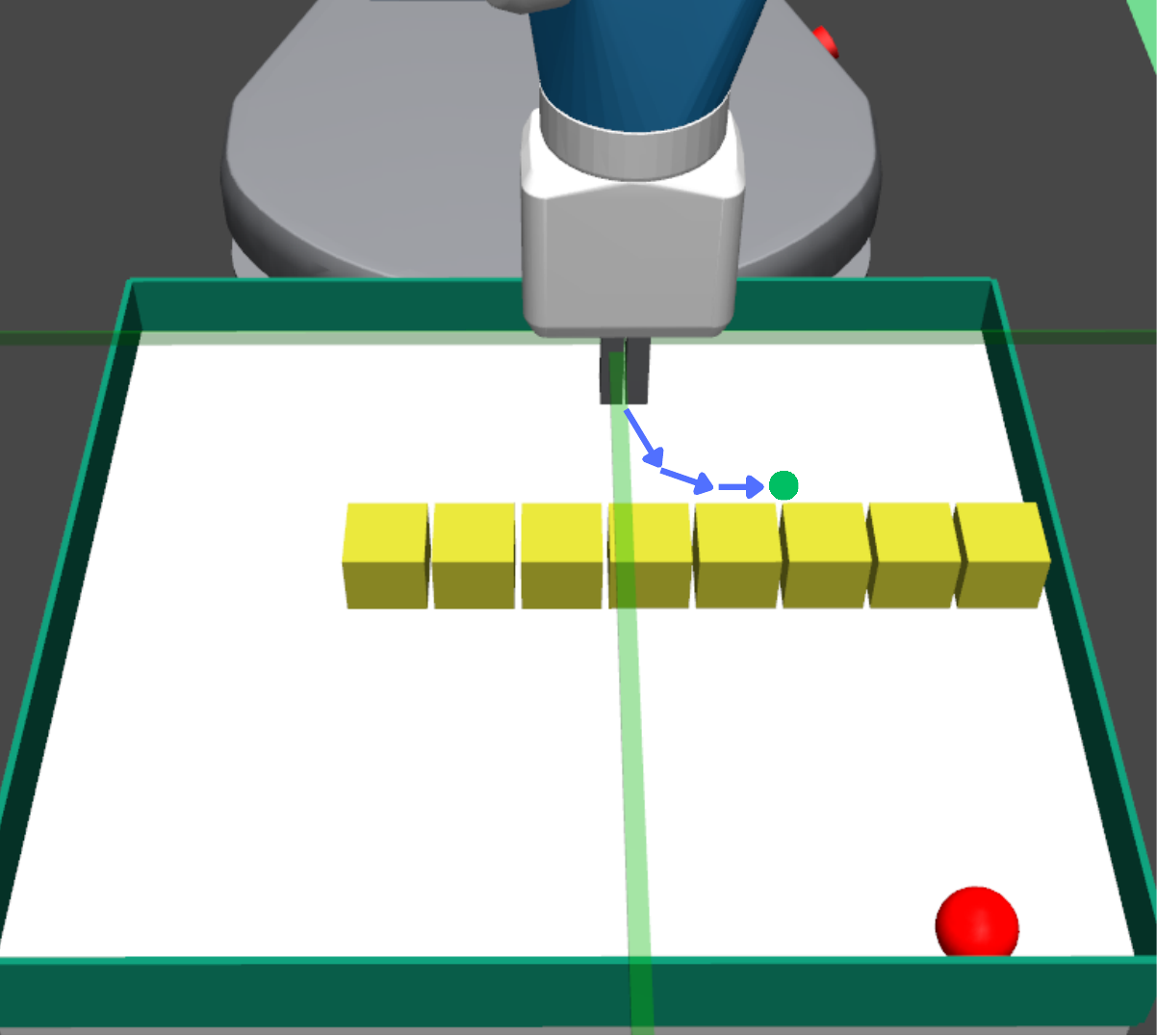}}
\hspace{0.03cm}
\subfloat[][VGRS (Ours)]{\includegraphics[scale=0.2]{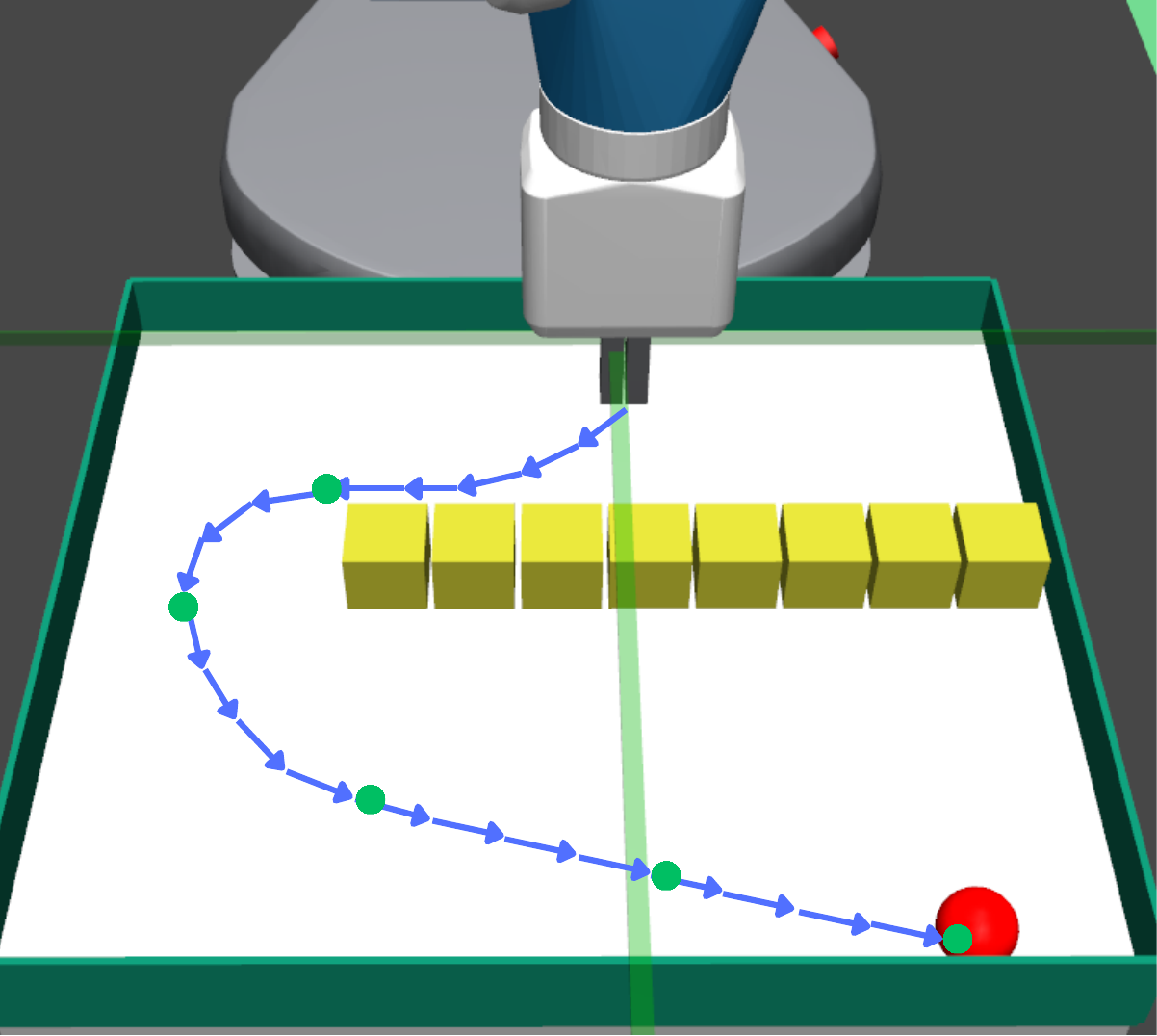}}
\\

\caption{\textbf{Visual feedback overcomes spatial local minima.} \textit{(Left)} An LLM-only static reward naively minimizes $L_2$ distance to the goal, trapping the agent on the right side of the wall. \textit{(Right)} VGRS leverages a VLM critic to diagnose the blockage. The LLM dynamically injects a region-specific penalty, successfully routing the agent through the left-side opening to reach the target.}
\label{fig:vlm_over_static}
\end{figure}

\subsection{RQ4: Qualitative Analysis of Visual Feedback} 
To highlight the necessity of the visual critic, we present a critical case study comparing our full VGRS pipeline against an ablation where the LLM was queried only once to generate a static, fixed reward function. For this, we designed a specific maze navigation scenario where the robot starts at the top-middle of the table and must navigate to a target at the bottom-right. The direct path is obstructed by a horizontal wall containing an opening exclusively on the far left. 

When relying solely on the text instruction, the static LLM baseline synthesized a naive reward function that simply minimized the $L_2$ distance between the robotic gripper and the goal. Because this reward formulation was visually blind to the obstacle topology, the policy repeatedly fell into a severe local minimum, driving the agent directly into the right side of the wall (Figure~\ref{fig:vlm_over_static}). Although the policy could eventually reach the goal through exhaustive random exploration, we found this approach to be highly sample-inefficient and prone to getting stuck in early rollouts.

In contrast, our VGRS framework successfully resolved this bottleneck. When training stagnated, the VLM analyzed the failed rollout frames, correctly detected the left-side opening, and explicitly provided this geometric feedback to the LLM. Leveraging this textual context, the LLM dynamically rewrote the reward code to inject a region-specific penalty whenever the agent proposed subgoals near the right side of the wall. This targeted penalty effectively forced the policy to explore leftward, successfully crossing the left wall opening to reach the final goal position. This shows that VGRS can systematically solve complex spatial bottlenecks that remain highly difficult for purely text-driven, static reward formulations to navigate. The synthesized reward code for both the static LLM-only ablation and our dynamic VGRS framework are provided in Appendix~\ref{appendix:case_study}.

\subsection{RQ5: How does VGRS compare with direct foundation model control at deployment time?}
To characterize the deployment trade-off in our target environment, we compare VGRS with representative foundation-model controllers, by evaluating on three direct-control baselines.
First, we deploy an off-the-shelf Vision-Language-Action model (\textit{SmolVLA}~\cite{shukor2025smolvla}). Because VLAs are tightly coupled to the kinematics of their pre-training data, zero-shot deployment on our 7-DoF Fetch environments results in severe action-space misalignment. Even when we implemented a custom heuristic controller to manually map the environment's observations and outputs to match the VLA's pre-training distribution, the model still failed to solve the task. Figure~\ref{fig:vgrs_comparison} summarizes the resulting behavior on the pick-and-place task. A VLA~\cite{shukor2025smolvla} fine-tuned on 100 expert demonstrations achieves a 29\% success rate, but its inference frequency remains at 50 Hz because the VLA backbone must be queried at every control step. In contrast, the VGRS policy achieves 81\% success, while running at approximately 500 Hz, since the foundation models are confined to the training phase and the deployed policy is lightweight.

Second, we evaluate a \textit{Direct-VLM Planner}, where a VLM processes visual frames to directly output high-level spatial waypoints. This baseline also failed, because it lacked the precise 3D geometric consistency required to generate stable, continuous coordinates, resulting in erratic, oscillatory trajectories.
Finally, we implement a \textit{Code as Policies (CaP)~\cite{liang2023code}} variant, prompting an LLM (\texttt{llama-3.3-70b}) to generate high-frequency end-effector offsets ($[\Delta x, \Delta y, \Delta z, \text{g}]$) at every step, where g defines whether the gripper is open (g=1) or closed (g=0). This approach similarly fails to complete the tasks; without closed-loop visual grounding or a learned low-level primitive to smooth the trajectory, the LLM-generated offsets accumulate massive compounding errors.
These results show that the considered baselines are unreliable under the robot and action-interface shift considered in our experiments. Fine-tuning partially improves VLA performance but retains the cost of querying the large backbone at every control step. By restricting the LLM and VLM strictly to synthesizing high-level rewards, VGRS delegates the complex embodied execution to the RL agent.

\subsection{RQ6: Sim-to-Real Transfer} 

\label{real_subsection}
To demonstrate the practical viability of our approach, we deploy VGRS on real-world pick-and-place and bin environments using a real Dobot Magician robotic arm equipped with a Realsense D435 depth camera. The learned policy evaluates at approximately 500 Hz on our inference benchmark. On the physical platform, the end-to-end control loop is capped at 20 Hz ($\Delta t = 50$ ms) by the manufacturer's SDK. We first train the hierarchical policies in simulation using the synthesized rewards, then evaluate the learned policy on the physical robot. We perform 5 sets of experiments with 10 trials each. VGRS achieves an average success rate of $60\%$ (variance $0.07$) on pick-and-place, and $60\%$ (variance $0.03$) on the bin task. We also deployed the baseline, L2R, which failed to complete the tasks on the physical hardware.
\section{DISCUSSION}
\textbf{Limitations.}
\label{sec:limitations}
While our approach VGRS effectively bridges the visual grounding gap without introducing inference-time latency, our event-triggered visual feedback loop assumes the VLM can accurately diagnose failure modes from a sparse sequence of images. In highly occluded manipulation tasks or tasks requiring fine tactile understanding (e.g., assessing the tightness of a grasp), the VLM's visual diagnosis may be insufficient. Furthermore, while restricting foundation models to the training phase ensures high-frequency control at deployment, it renders the final policy unable to dynamically adapt to entirely new, zero-shot language instructions without initiating a new training cycle. Although such a retraining process is computationally lightweight, improving the zero-shot generalization of these compiled policies to novel semantic instructions remains a highly promising direction for future work.

\textbf{Conclusion.}
\label{sec:conclusion}
We propose Visually-Grounded Reward Synthesis (VGRS), a framework that shifts foundation models from deployment-time control to training-time supervision, enabling lightweight hierarchical policies that execute with low latency at deployment. VGRS employs a decoupled Vision-Language-Model (VLM) and Large-Language-Model (LLM) pipeline to translate natural-language task descriptions into executable reward functions. When policy learning stagnates, an event-triggered visual critic analyzes failed rollouts to identify the underlying physical failure mode, and the LLM uses this diagnosis to iteratively revise and densify the reward. Unlike existing Language-to-Reward methods, which generate rewards without observing how the policy executes, VGRS closes the loop between policy execution and reward design. Our experiments on simulated and real-world tasks demonstrate that VGRS substantially improves long-horizon task success while requiring only the learned lightweight policy during deployment.

\bibliography{IEEEexample}
\bibliographystyle{IEEEtran}

\onecolumn
\appendix
\section{APPENDIX}
\tableofcontents
\setlength{\parindent}{0pt}

\subsection{VGRS: Pseudo-Code}
\label{appendix:algorithm}
\begin{algorithm}[H]
\caption{Visually-Grounded Reward Synthesis (VGRS)}
\label{alg:vgrs}
\begin{algorithmic}[1]
\Require Task instruction $\mathcal{T}$, Environment APIs $\mathcal{P}_{\text{env}}$, Pre-trained LLM $\Phi_{\text{LLM}}$, Pre-trained VLM $\Psi_{\text{VLM}}$, High-level policy $\pi_H$, Low-level policy $\pi_L$, Replay buffers $\mathcal{D}_H, \mathcal{D}_L$, Primitive horizon $k$, Stagnation threshold $\eta_{\text{threshold}}$

\State \textbf{// Phase 1: Initial Reward Synthesis}
\State $R_H^{(0)} \leftarrow \Phi_{\text{LLM}}(\mathcal{T}, \mathcal{P}_{\text{env}})$ \Comment{Zero-shot reward generation}
\State $m \leftarrow 0$ \Comment{Initialize reward version index}
\State Initialize success rate moving average $\eta_w \leftarrow 0$

\State \textbf{// Phase 2 \& 3: Hierarchical Optimization \& Visual Feedback}
\For{episode $= 1$ to $N$}
    \State Reset environment, obtain initial state $s_0$
    
    \For{step $t = 0, k, 2k, \dots, T-k$}
        \State $g_t \sim \pi_H(\cdot \mid s_t)$ \Comment{Propose high-level subgoal}
        
        \For{primitive step $i = 0$ to $k-1$}
            \State $a_{t+i} \sim \pi_L(\cdot \mid s_{t+i}, g_t)$ \Comment{Sample low-level action}
            \State Execute $a_{t+i}$, observe next state $s_{t+i+1}$
            \State $R_L \leftarrow -\|s_{t+i+1} - g_t\|_2$
            \State Store low-level transition $(s_{t+i}, g_t, a_{t+i}, R_L, s_{t+i+1})$ in $\mathcal{D}_L$
        \EndFor
        
        \State Store high-level transition $(s_t, g^\star, g_t, s_{t+k})$ in $\mathcal{D}_H$
        
        \State \textbf{// Concurrent Policy Updates}
        \State Sample minibatch from $\mathcal{D}_L$ and update $\pi_L$ via SAC
        \State Sample minibatch from $\mathcal{D}_H$ 
        \State Relabel batch rewards using active reward: $r_H = R_H^{(m)}(s_{t+k}, g_t, g^\star)$
        \State Update $\pi_H$ via SAC using the relabeled transitions
    \EndFor
    
    \State Update success rate moving average $\eta_w$
    
    \State \textbf{// Event-Triggered Visual Feedback}
    \If{$\nabla \eta_w \approx 0$ \textbf{and} $\eta_w < \eta_{\text{threshold}}$}
        \State $I_{\text{fail}} \leftarrow$ Extract visual frames from the most recent failed trajectory
        \State $\mathcal{F} \leftarrow \Psi_{\text{VLM}}(I_{\text{fail}}, \mathcal{T})$ \Comment{VLM diagnoses physical bottleneck}
        \State $R_H^{(m+1)} \leftarrow \Phi_{\text{LLM}}(\mathcal{T}, \mathcal{F}, R_H^{(m)})$ \Comment{LLM synthesizes dense reward}
        \State $m \leftarrow m + 1$ \Comment{Increment reward version}
        \State Reset $\eta_w$ to prevent immediate re-triggering
    \EndIf
\EndFor
\end{algorithmic}
\end{algorithm}

\subsection{Implementation details}
\label{sec:implementation_details}
In our setup, both the actor and critic networks are implemented as three-layer, fully connected neural networks, each with 512 neurons per layer.

For the maze navigation task, a 7-degree-of-freedom (7-DoF) robotic arm moves through a four-room maze with its closed gripper fixed at table height, navigating to reach the goal position. In the pick-and-place task, the same 7-DoF robotic arm identifies a square block, picks it up, and delivers it to the goal position. In the bin environment, the gripper must pick up the block and place it in a designated bin. Lastly, in the kitchen task, a 9-DoF Franka robot performs a predefined complex action—opening a microwave door—to complete the task.

To ensure fair comparisons, we maintain consistency across all baselines by keeping key parameters unchanged wherever possible. These include the neural network layer width, the number of layers, the choice of optimizer, and the SAC implementation parameters. We provide the hyperparameter configuration in Table~\ref{tab:hierarchical_hyperparams}

\begin{table}[h]
\centering
\caption{Hyperparameter Configuration}
\begin{tabular}{|l|c|l|}
\hline
\textbf{Parameter} & \textbf{Value} & \textbf{Description} \\
\hline
activation & tanh & activation for hierarchical policies \\
layers & 3 & number of layers in the critic/actor networks \\
hidden & 512 & number of neurons in each hidden layer \\
Q\_lr & 0.001 & critic learning rate \\
pi\_lr & 0.001 & actor learning rate \\
buffer\_size & int(1E7) & for experience replay \\
tau & 0.8 & polyak averaging coefficient \\
clip\_obs & 200 & clip observation \\
n\_cycles & 1 & per epoch \\
n\_batches & 10 & training batches per cycle \\
batch\_size & 1024 & batch size hyper-parameter \\
random\_eps & 0.2 & percentage of time a random action is taken \\
alpha & 0.05 & weightage parameter for SAC \\
noise\_eps & 0.05 & std of gaussian noise added to not-completely-random actions \\
norm\_eps & 0.01 & epsilon used for observation normalization \\
norm\_clip & 5 & normalized observations are cropped to this value \\
adam\_beta1 & 0.9 & beta 1 for Adam optimizer \\
adam\_beta2 & 0.999 & beta 2 for Adam optimizer \\
\hline
\end{tabular}
\label{tab:hierarchical_hyperparams}
\end{table}

\subsection{Environment details}
\label{sec:environment_details}
In this section, we provide the environment and implementation details for all the tasks:

\subsubsection{Maze Navigation Environment} 

In this environment, a $7$-DOF robotic arm gripper navigates through randomly generated four-room mazes to reach the goal position. The gripper remains closed and fixed at table height, with the positions of walls and gates randomly determined. The table is divided into a rectangular $W \times H$ grid, and the vertical and horizontal wall positions, $W_{P}$ and $H_{P}$, are randomly selected from $(1, W-2)$ and $(1, H-2)$, respectively. In the constructed four-room environment, the four gate positions are randomly chosen from $(1, W_{P}-1)$, $(W_{P}+1, W-2)$, $(1, H_{P}-1)$, and $(H_{P}+1, H-2)$.

\par In the maze environment, the state is represented as the vector $[dx, M]$, where $dx$ denotes the current gripper position and $M$ is the sparse maze array. The higher-level policy input is a concatenated vector $[dx, M, g]$, where $g$ is the target goal position. The lower-level policy input is a concatenated vector $[dx, M, s_g]$, where $s_g$ is the sub-goal provided by the higher-level policy. $M$ is a discrete 2D one-hot vector array, with $1$ indicating the presence of a wall block. The lower primitive action $a$ is a 4-dimensional vector, with each dimension $a_i \in [0, 1]$. The first three dimensions provide offsets to be scaled and added to the gripper position for movement. The last dimension controls the gripper, with $0$ indicating a closed gripper and $1$ indicating an open gripper.

\subsubsection{Pick and Place and Bin Environments} 

In this section, we describe the environment details for the pick and place and bin tasks. The state is represented as the vector $[dx, o, q, e]$, where $dx$ is the current gripper position, $o$ is the position of the block object on the table, $q$ is the relative position of the block with respect to the gripper, and $e$ includes the linear and angular velocities of both the gripper and the block object. The higher-level policy input is a concatenated vector $[dx, o, q, e, g]$, where $g$ is the target goal position. The lower-level policy input is a concatenated vector $[dx, o, q, e, s_g]$, where $s_g$ is the sub-goal provided by the higher-level policy. In our experiments, the sizes of $dx$, $o$, $q$, and $e$ are set to $3$, $3$, $3$, and $11$, respectively. The lower primitive action $a$ is a 4-dimensional vector with each dimension $a_i \in [0, 1]$. The first three dimensions provide gripper position offsets, and the last dimension controls the gripper. During training, the positions of the block object and the goal are randomly generated (the block is always initialized on the table, and the goal is always above the table at a fixed height). 

\subsubsection{Franka Kitchen Environment} 
For this environment please refer to the D4RL environment~\cite{DBLP:journals/corr/abs-2004-07219}. In this environment, the franka robot has to perform a complex multi-stage task in order to achieve the final goal.

\subsection{RQ3: Comparing our VLM+LLM pipeline against VLM only baseline for reward code generation}
\label{appendix:vlm_only}
As seen in these examples, our VLM+LLM pipeline generates correct code, whereas VLM only pipeline generates incorrect code, by hallucinating block positions and making frequent dimensionality mistakes.

\begin{mdframed}[backgroundcolor=white]
\vspace{1em}
\noindent \textbf{VGRS (VLM+LLM Example)): A code generation example in pick and place task using VLM+LLM pipeline.}
\vspace{1em}

\noindent \textbf{[VLM Critic Diagnosis]:} The robot failed to navigate to the bottom right corner of the table because it got stuck behind the wall of blocks in the middle of the table. The robot is consistently seen to the left or top of the block wall structure throughout the sequence, never reaching the bottom right corner where the red ball is located.

\vspace{1em}
\noindent \textbf{[LLM Coder]} Synthesizing new grounded DENSE reward...\\

\vspace{1em}
\begin{verbatim}
import numpy as np

def compute_reward_llm(self, achieved_goal, goal=None):
    distance_threshold = 0.05
    d_threshold = 0.1
    if goal is None:
        x_index = 9 
        y_index = 9
        goal = self.get_goal_pos_llm(x_index, y_index)
        goal = np.array(goal).flatten()
        
    d = goal_distance_llm(achieved_goal, goal)
    sparse_reward = -(d > distance_threshold).astype(np.float32)
    
    # dense penalty for being stuck behind the wall of blocks
    block_wall_x = 5  # assuming the block wall is at x=5
    block_wall_y = 5  # assuming the block wall is at y=5
    if achieved_goal[0] < block_wall_x and achieved_goal[1] < block_wall_y:
        dense_reward = -d
    else:
        dense_reward = -np.abs(achieved_goal[0] - goal[0]) - \
                       np.abs(achieved_goal[1] - goal[1])
        
    dense_reward *= d_threshold
    return sparse_reward + dense_reward

def goal_distance_llm(goal_a, goal_b):
    assert goal_a.shape == goal_b.shape
    return np.linalg.norm(goal_a - goal_b, axis=-1)
\end{verbatim}
\end{mdframed}


\begin{mdframed}[backgroundcolor=white]
\textbf{VLM Only Baseline Example: A code generation example in pick and place task using VLM only pipeline.}

\vspace{1em}
\noindent \textbf{[VLM Coder]} Synthesizing new grounded DENSE reward...
\vspace{1em}
\begin{verbatim}
import numpy as np

def compute_reward_llm(self, achieved_goal, goal=None):
    # Sparse reward for achieving the goal
    sparse_reward = 0.0
    if np.all(achieved_goal == goal):
        sparse_reward = 1.0

    # Dense reward for penalizing block displacement and out-of-bounds movement
    goal_distance_llm = goal[:len(achieved_goal)]
    d = np.abs(achieved_goal - goal_distance_llm)
    dense_reward = -0.5 * np.mean(d)  # Penalty for displacement

    # Additional penalty for blocks out of bounds or misplaced
    block_out_of_bounds_penalty = 0.0
    for block in achieved_goal:
        # Assuming board bounds are [0,1] for both x and y
        if block[0] < 0 or block[0] > 1 or block[1] < 0 or block[1] > 1: 
            block_out_of_bounds_penalty -= 0.1  # Penalty for block out of bounds

    misplaced_blocks_penalty = 0.0
    for i in range(len(achieved_goal)):
        # Check if block is close to goal position
        if not np.allclose(achieved_goal[i], goal_distance_llm[i], atol=0.1):  
            misplaced_blocks_penalty -= 0.05  # Penalty for each misplaced block

    dense_reward += block_out_of_bounds_penalty + misplaced_blocks_penalty

    # Ensure dense reward is within [-1.0, 1.0]
    dense_reward = np.clip(dense_reward, -1.0, 1.0)

    # Apply threshold and combine with sparse reward
    d_threshold = 0.1
    dense_reward = dense_reward * d_threshold

    total_reward = sparse_reward + dense_reward

    return total_reward
\end{verbatim}
\end{mdframed}

\subsection{RQ4: Qualitative Case Study of LLM vs. VGRS Reward Synthesis}
\label{appendix:case_study}
\begin{mdframed}[backgroundcolor=white]

\textbf{LLM Static Code Generation Example: Below is the initial reward code generated by the LLM relying strictly on the text instruction, which resulted in the agent getting stuck in a local minimum on the right side of the wall.}
\vspace{1em}
\begin{verbatim}
import numpy as np

def compute_reward_llm(self, achieved_goal, goal=None):
    distance_threshold = 0.05
    d_threshold = 0.1
    if goal is None:
        x_index = 9 
        y_index = 9
        goal = self.get_goal_pos_llm(x_index, y_index)
        goal = np.array(goal).flatten()
        
    d = goal_distance_llm(achieved_goal, goal)
    sparse_reward = -(d > distance_threshold).astype(np.float32)
    
    return sparse_reward

def goal_distance_llm(goal_a, goal_b):
    assert goal_a.shape == goal_b.shape
    return np.linalg.norm(goal_a - goal_b, axis=-1)
\end{verbatim}
\end{mdframed}

\begin{mdframed}[backgroundcolor=white]
\textbf{VGRS Dynamic Code Generation Example: Here we present the case study, where after the VLM diagnosed the failure and provided feedback that the wall is open on the left, the LLM dynamically rewrote the reward code. Note the addition of the region-specific penalty to force leftward exploration.}

\vspace{1em}

\textbf{[VLM Critic Diagnosis]:} The robot failed to navigate to the bottom right corner of the table because it got stuck behind the wall of blocks in the middle of the table. The robot is consistently seen to the left or top of the block wall structure throughout the sequence, never reaching the bottom right corner where the red ball is located.

\vspace{1em}
\textbf{[LLM Coder]} Synthesizing new grounded DENSE reward...\\

\begin{verbatim}
import numpy as np

def compute_reward_llm(self, achieved_goal, goal=None):
    distance_threshold = 0.05
    d_threshold = 0.1
    if goal is None:
        x_index = 9 
        y_index = 9
        goal = self.get_goal_pos_llm(x_index, y_index)
        goal = np.array(goal).flatten()
        
    d = goal_distance_llm(achieved_goal, goal)
    sparse_reward = -(d > distance_threshold).astype(np.float32)
    
    # dense penalty for being stuck behind the wall of blocks
    block_wall_y = 1  # assuming the left wall is at y=1
    dense_reward = -np.abs(achieved_goal[1] - block_wall_y)
    
    dense_reward *= d_threshold
    return sparse_reward + dense_reward

def goal_distance_llm(goal_a, goal_b):
    assert goal_a.shape == goal_b.shape
    return np.linalg.norm(goal_a - goal_b, axis=-1)
\end{verbatim}
\end{mdframed}

\subsection{Reward Synthesizers Prompts and VLM Critic Prompts}
\label{appendix:templates}
\textbf{1. Direct Geometric Reward Synthesizer and VLM Critic Prompt for Maze Navigation Environment}
\vspace{0.5em}

\begin{mdframed}[backgroundcolor=white]
\textbf{[Stage 1: Initial LLM Reward Generation]}\\
You are an expert Python robotics programmer.\\
Write a python function named \texttt{compute\_reward\_llm} that will be bound to a reinforcement learning environment class.

\textbf{[Task Description]}\\
'Navigate to the bottom right corner of the table.'

\textbf{[Environment Coordinate Logic]}\\
The table grid is (10,10). X indices are 0 to 9. Y indices are 0 to 9.\\
Top left is (0,0). Bottom right is (9,9).\\
There are four rooms. The robot starts from the top left room. There are four doors, top, left, right and bottom.\\

\textbf{[Stage 2: VLM Visual Critic Observation]}\\
\textit{System Input (Images):} [Frames of the failed rollout]\\
\textit{System Prompt:} "You are a roboticist observing a failed rollout for the task: 'Navigate to the bottom right corner'. Mention the room the robot is currently in. Describe purely physically exactly where/why it failed. Be specific about spatial relations (e.g., 'got stuck at the first door'). Keep it to 2 sentences."

[VLM Output]

\textbf{[Stage 3: LLM Dense Reward Refinement]}\\
The current Python reward function is: [Sparse Code]\\
A Vision-Language Model observed the robot failing and reported:

[VLM Output]

Available variables inside the environment class: \texttt{self.door\_positions}: list of [x,y] coordinates of the maze doors.\\
You must REWRITE \texttt{compute\_reward\_llm} to add a DENSE continuous penalty heuristic that explicitly addresses the VLM's observation. Multiply the dense\_reward with d\_threshold = 0.1 before adding to the sparse reward.
\end{mdframed}
\vspace{1em}

\textbf{2. Direct Geometric Reward Synthesizer and VLM Critic Prompt for Pick and Place}
\vspace{0.5em}

\begin{mdframed}[backgroundcolor=white]
\textbf{[Stage 1: Initial LLM Reward Generation]}\\
You are an expert Python robotics programmer.\\
Write a python function named \texttt{compute\_reward\_llm} that will be bound to a reinforcement learning environment class.

\textbf{[Task Description]}\\
'Pick the block and raise it above the table.'

\textbf{[Environment Coordinate Logic]}\\
The robot is a 7-DOF robotic arm. The table height is Z = 0.42.\\
The maximum reachable height for the gripper is Z = 0.66.\\

\textbf{[Stage 2: VLM Visual Critic Observation]}\\
\textit{System Input (Images):} [Frames of the failed rollout]\\
\textit{System Prompt:} "You are a roboticist observing a failed rollout for the task: 'Pick the block and raise it above the table'. Describe purely physically exactly where/why it failed. Be specific about spatial relations (e.g., 'the gripper completely missed the block', 'the block was dropped before reaching the target height'). Keep it to 2 sentences."

[VLM Output]

\textbf{[Stage 3: LLM Dense Reward Refinement]}\\
The current Python reward function is: [Sparse Code]\\
A Vision-Language Model observed the robot failing and reported:

[VLM Output]

You must REWRITE \texttt{compute\_reward\_llm} to add a DENSE continuous penalty heuristic that explicitly addresses the VLM's observation.
\end{mdframed}
\vspace{1em}






\textbf{3. Direct Geometric Reward Synthesizer and VLM Critic Prompt for Bin Environment}
\vspace{0.5em}

\begin{mdframed}[backgroundcolor=white]
\textbf{[Stage 1: Initial LLM Reward Generation]}\\
You are an expert Python robotics programmer.\\
Write a python function named \texttt{compute\_reward\_llm} that will be bound to a reinforcement learning environment class.

\textbf{[Task Description]}\\
'Pick the block and place it inside the bin.'

\textbf{[Environment Coordinate Logic]}\\
The robot is a 7-DOF robotic arm. The table height is Z = 0.42.\\
There is a bin on the table. The object must be placed at the continuous (x, y) center of the bin.\\

\textbf{[Stage 2: VLM Visual Critic Observation]}\\
\textit{System Input (Images):} [Frames of failed rollout]\\
\textit{System Prompt:} "You are a roboticist observing a failed rollout for the task: 'Pick the block and place it inside the bin'. Describe purely physically exactly where/why it failed. Be specific about spatial relations (e.g., 'the block fell outside the bin', 'the gripper failed to release the object'). Keep it to 2 sentences."

[VLM Output]

\textbf{[Stage 3: LLM Dense Reward Refinement]}\\
The current Python reward function is: [Sparse Code]\\
A Vision-Language Model observed the robot failing and reported:

[VLM Output]

Available variables inside the environment class: \texttt{self.bin\_position}: [x,y,z] coordinates of the bin.\\
You must REWRITE \texttt{compute\_reward\_llm} to add a DENSE continuous penalty heuristic that explicitly addresses the VLM's observation.
\end{mdframed}
\vspace{1em}

\textbf{4. Direct Geometric Reward Synthesizer and VLM Critic Prompt for Franka Kitchen}
\vspace{0.5em}

\begin{mdframed}[backgroundcolor=white]
\textbf{[Stage 1: Initial LLM Reward Generation]}\\
You are an expert Python robotics programmer.\\
Write a python function named \texttt{compute\_reward\_llm} that will be bound to a reinforcement learning environment class.

\textbf{[Task Description]}\\
'Open the microwave door fully, then turn the gas knob to the on position'

\textbf{[Environment Coordinate Logic]}\\
The robot is a 9-DOF Franka arm operating in a kitchen.\\
The task involves articulating objects, represented by joint states rather than spatial grid coordinates.\\
The microwave door is controlled by a single joint \texttt{joint\_1}. Fully closed is 0.0, fully open is -0.75.\\

\textbf{[Stage 2: VLM Visual Critic Observation]}\\
\textit{System Input (Images):} [Frames of failed rollout]\\
\textit{System Prompt:} "You are a roboticist observing a failed rollout for the task: 'Open the microwave door fully, then turn the gas knob'. Describe purely physically exactly where/why it failed. Be specific about spatial relations (e.g., 'the gripper grasped the wrong handle', 'the microwave door is only halfway open'). Keep it to 2 sentences."

[VLM Output]

\textbf{[Stage 3: LLM Dense Reward Refinement]}\\
The current Python reward function is: [Sparse Code]\\
A Vision-Language Model observed the robot failing and reported:

[VLM Output]

Available variables inside the environment class: \texttt{self.knob\_position}: [x,y,z] spatial coordinates of the target knob.\\
You must REWRITE \texttt{compute\_reward\_llm} to add a DENSE continuous penalty heuristic that explicitly addresses the VLM's observation to guide the end-effector closer to the knob.
\end{mdframed}








\subsection{Qualitative Visualizations}
\label{appendix:qualitative_vizualizations}
We provide qualitative visualizations for all the environments:

\begin{figure}[H]
\vspace{5pt}
\centering
\includegraphics[scale=0.09]{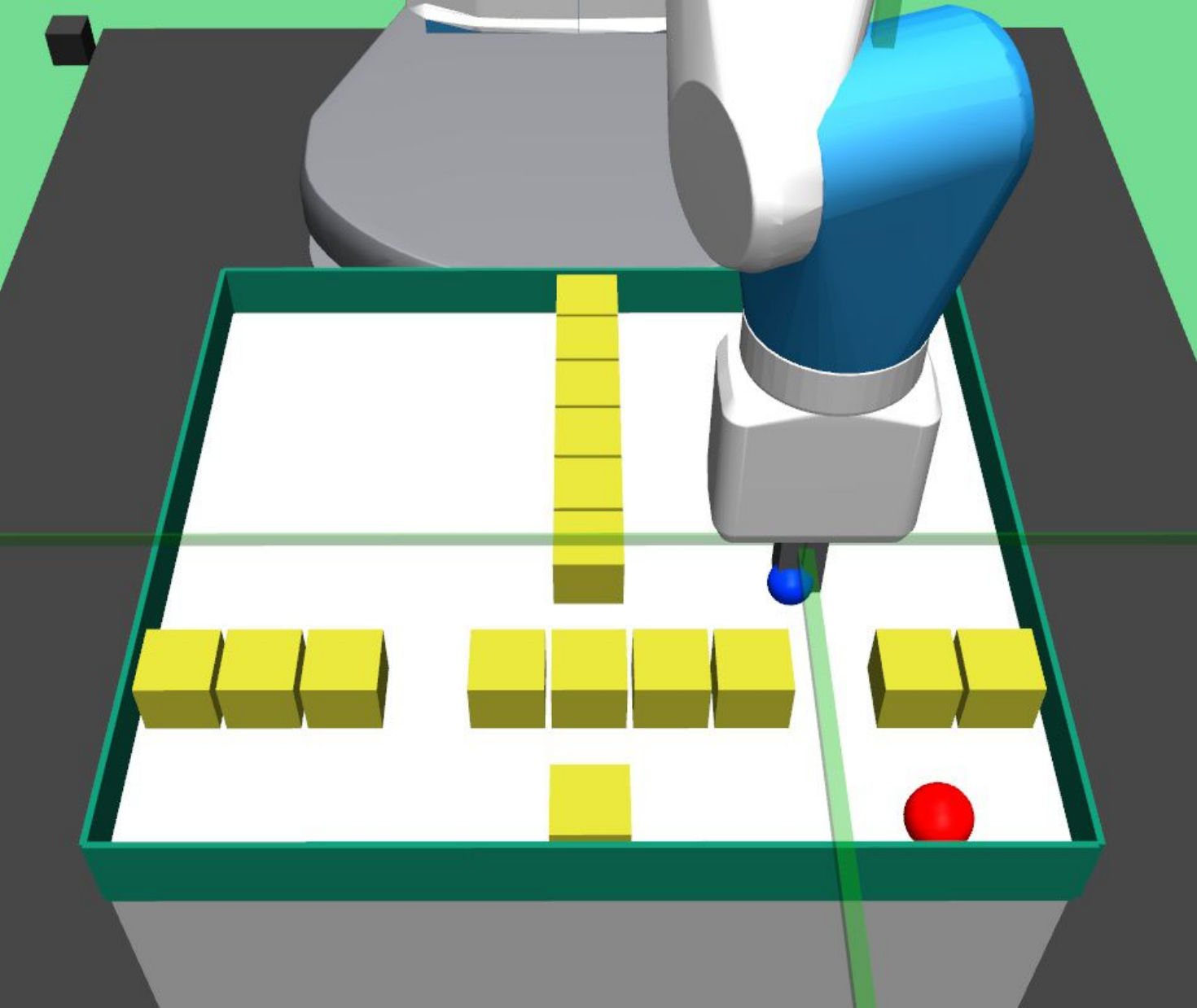}
\includegraphics[scale=0.09]{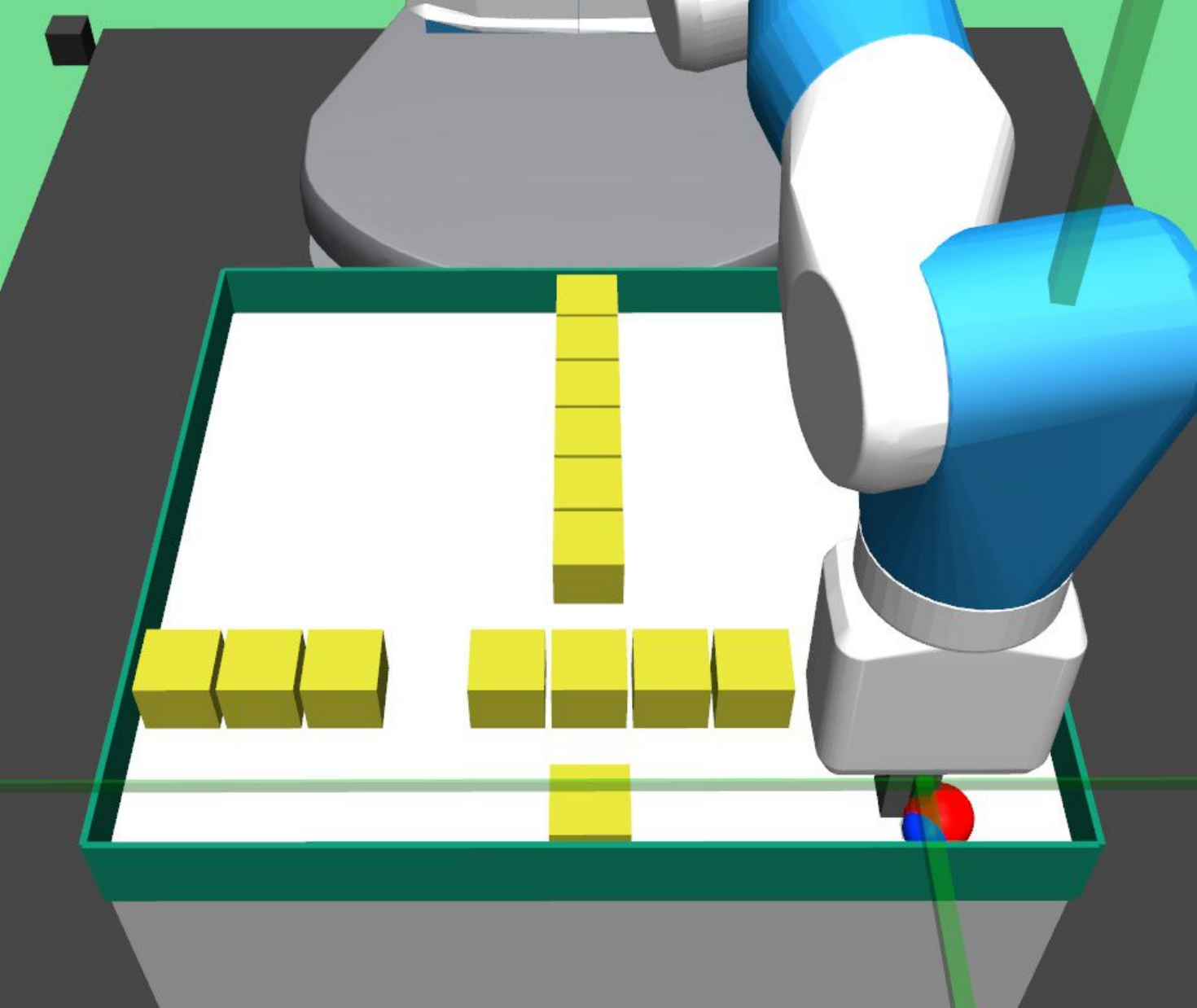}
\includegraphics[scale=0.09]{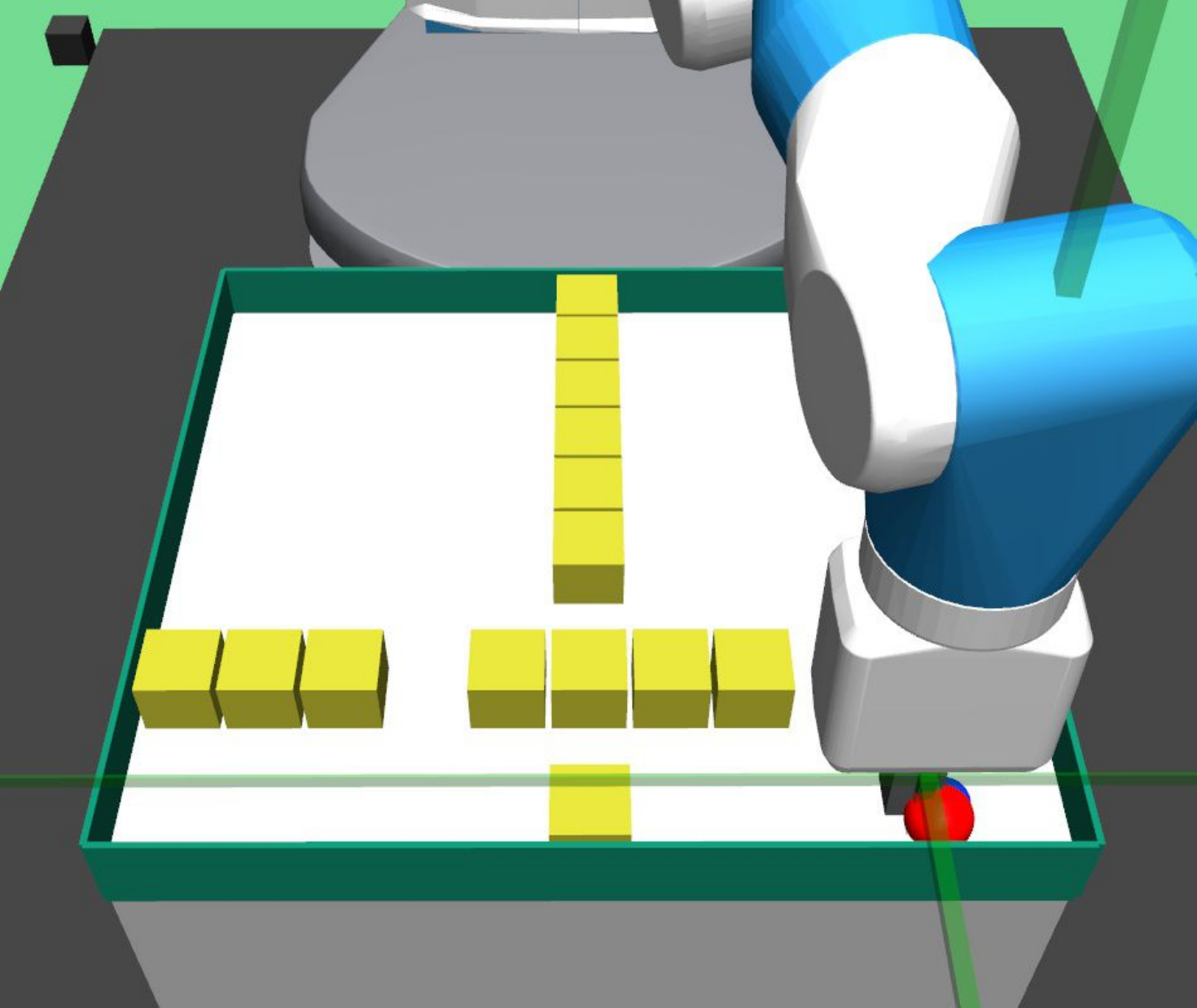}
\includegraphics[scale=0.09]{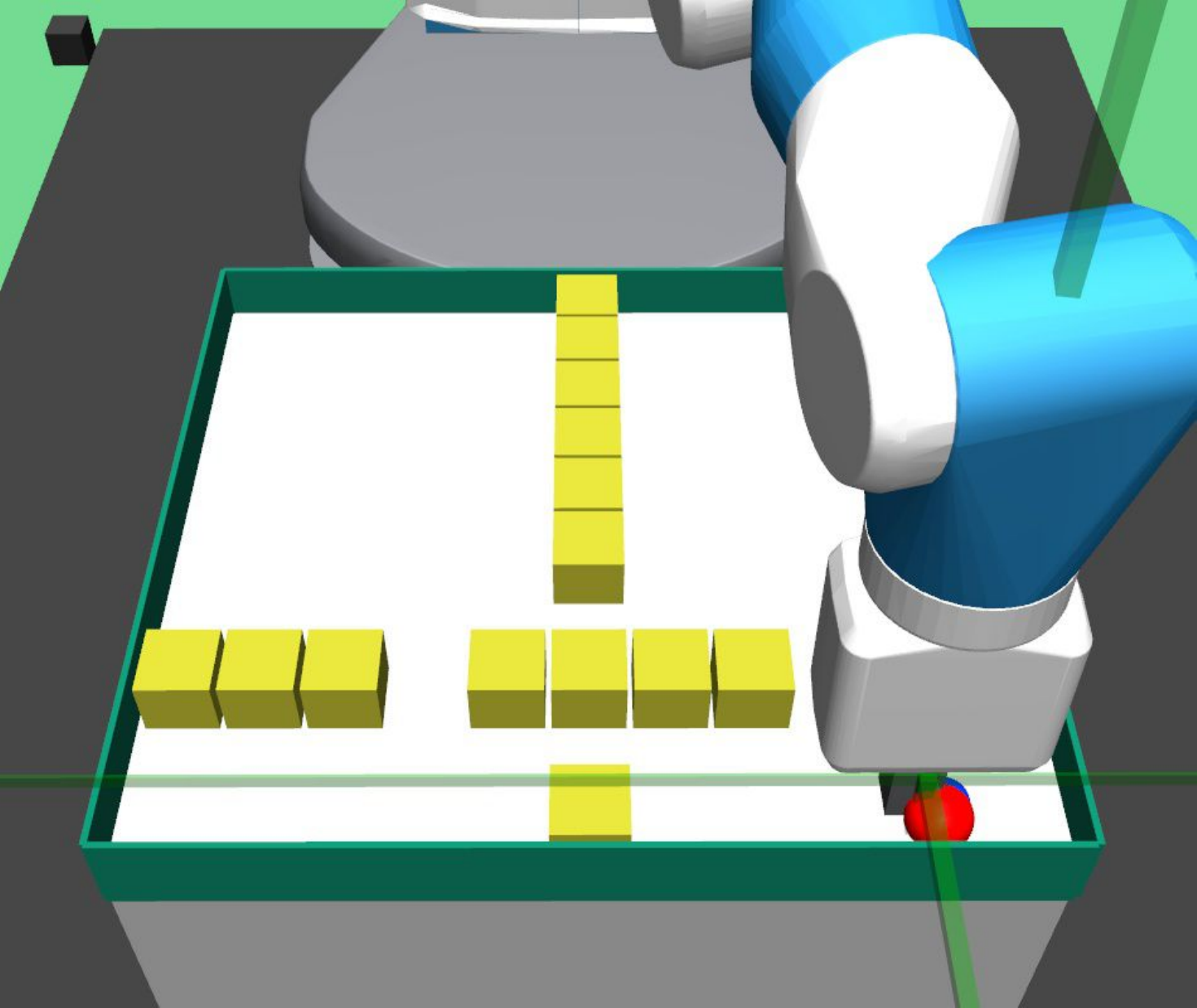}
\includegraphics[scale=0.09]{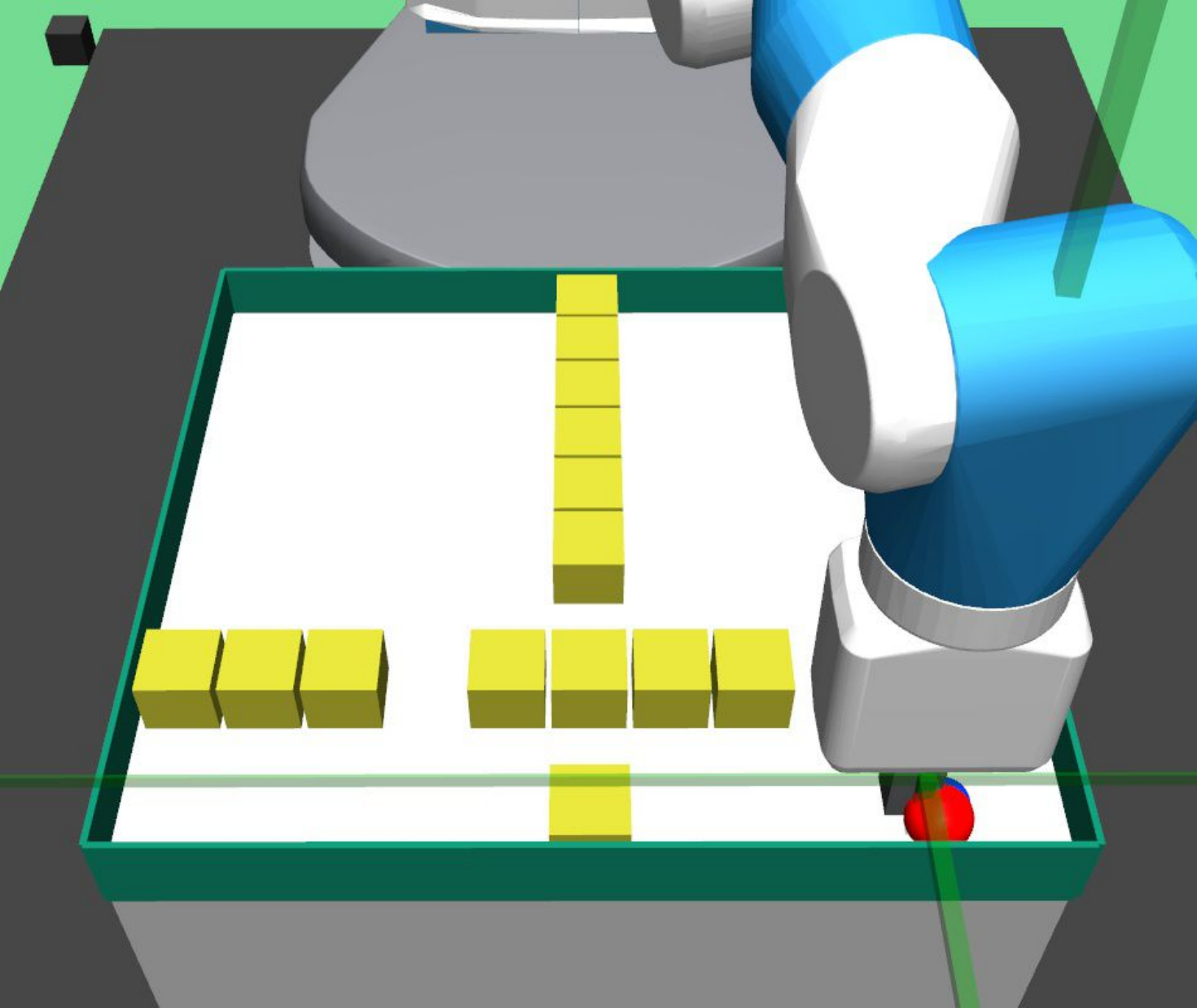}
\includegraphics[scale=0.09]{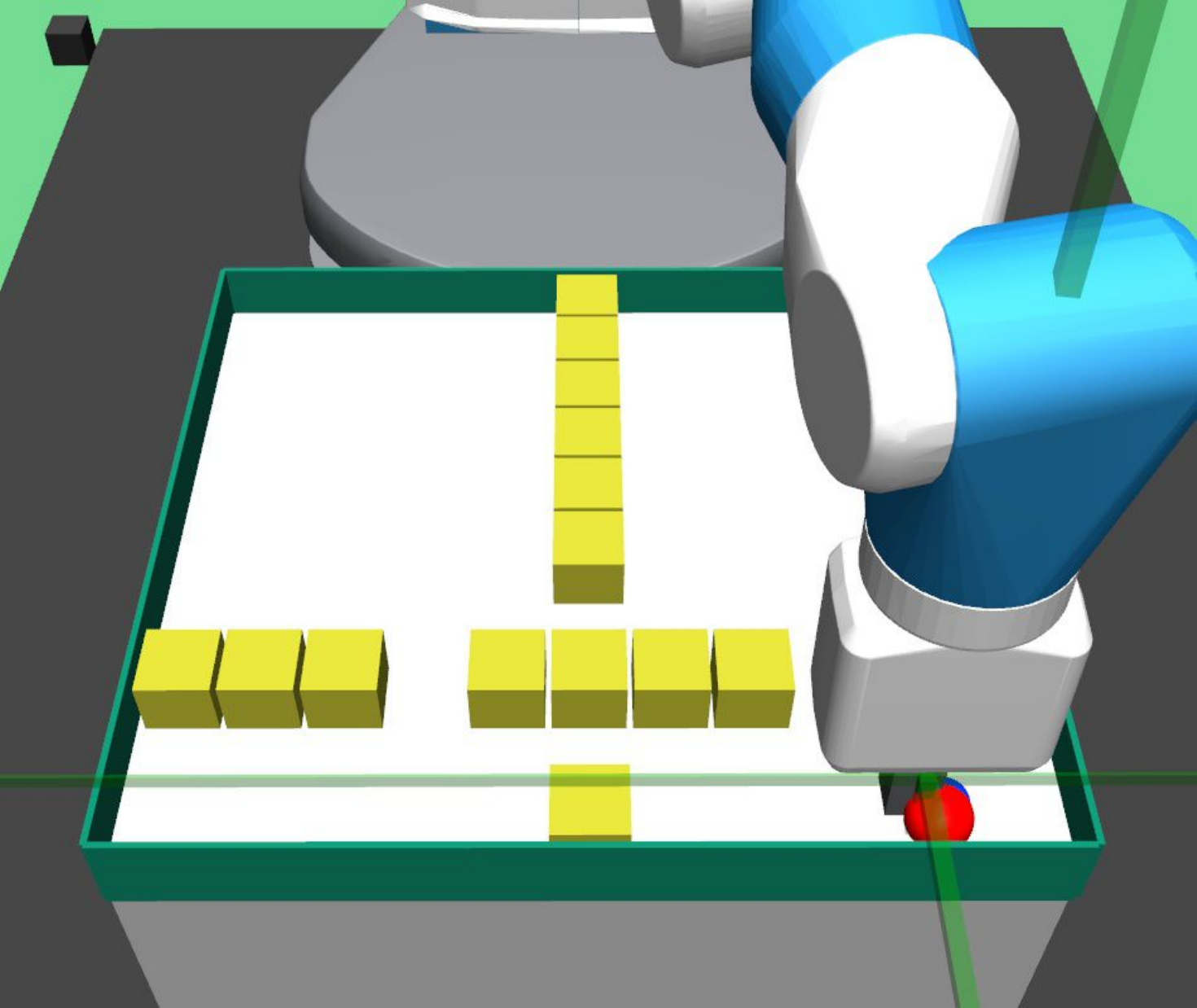}
\caption{\textbf{Successful visualization}: The visualization is a successful attempt at performing maze navigation task}
\label{fig:maze_viz_success_2_ablation.}
\end{figure}


\begin{figure}[ht]
\vspace{5pt}
\centering
\includegraphics[scale=0.09]{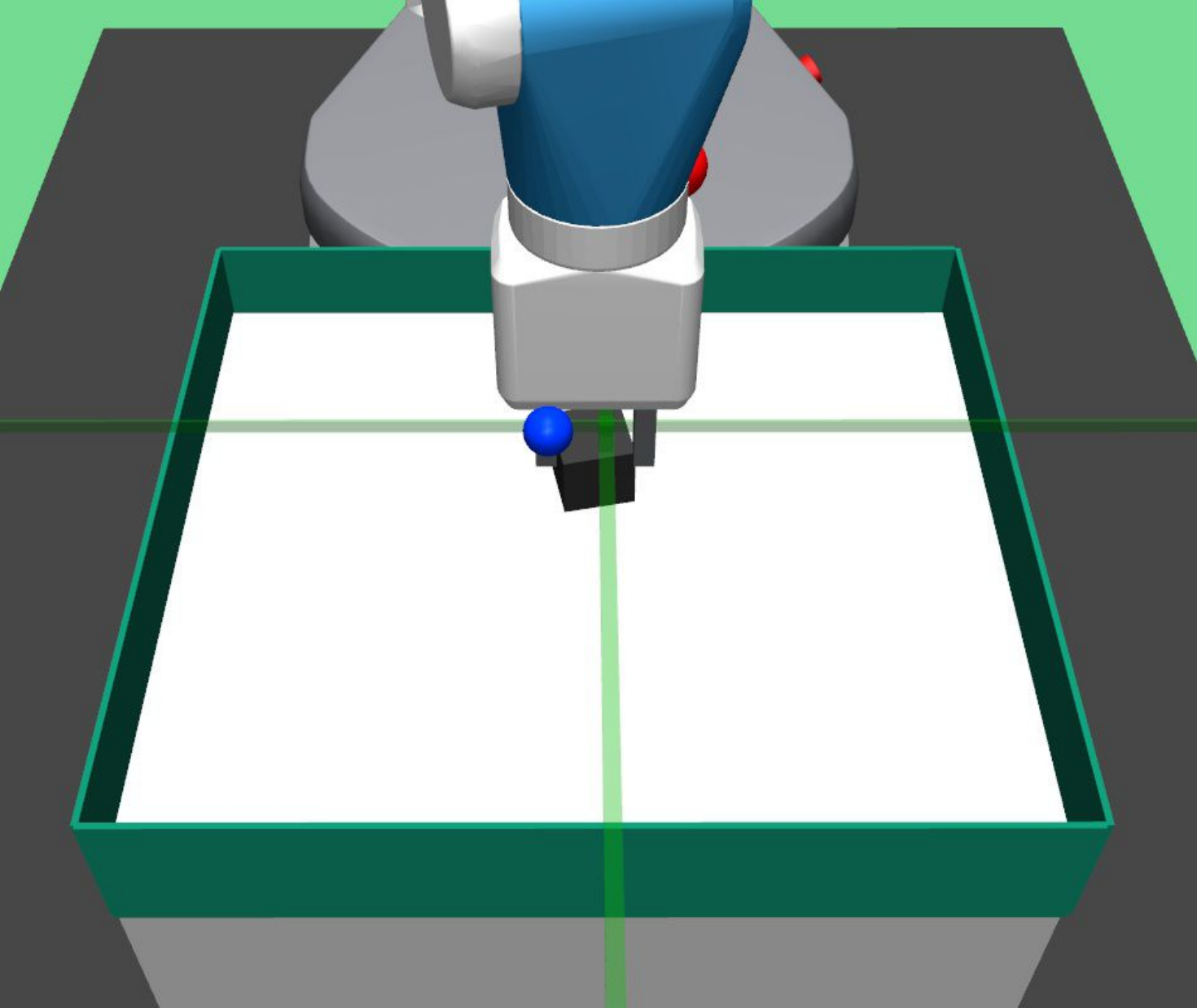}
\includegraphics[scale=0.09]{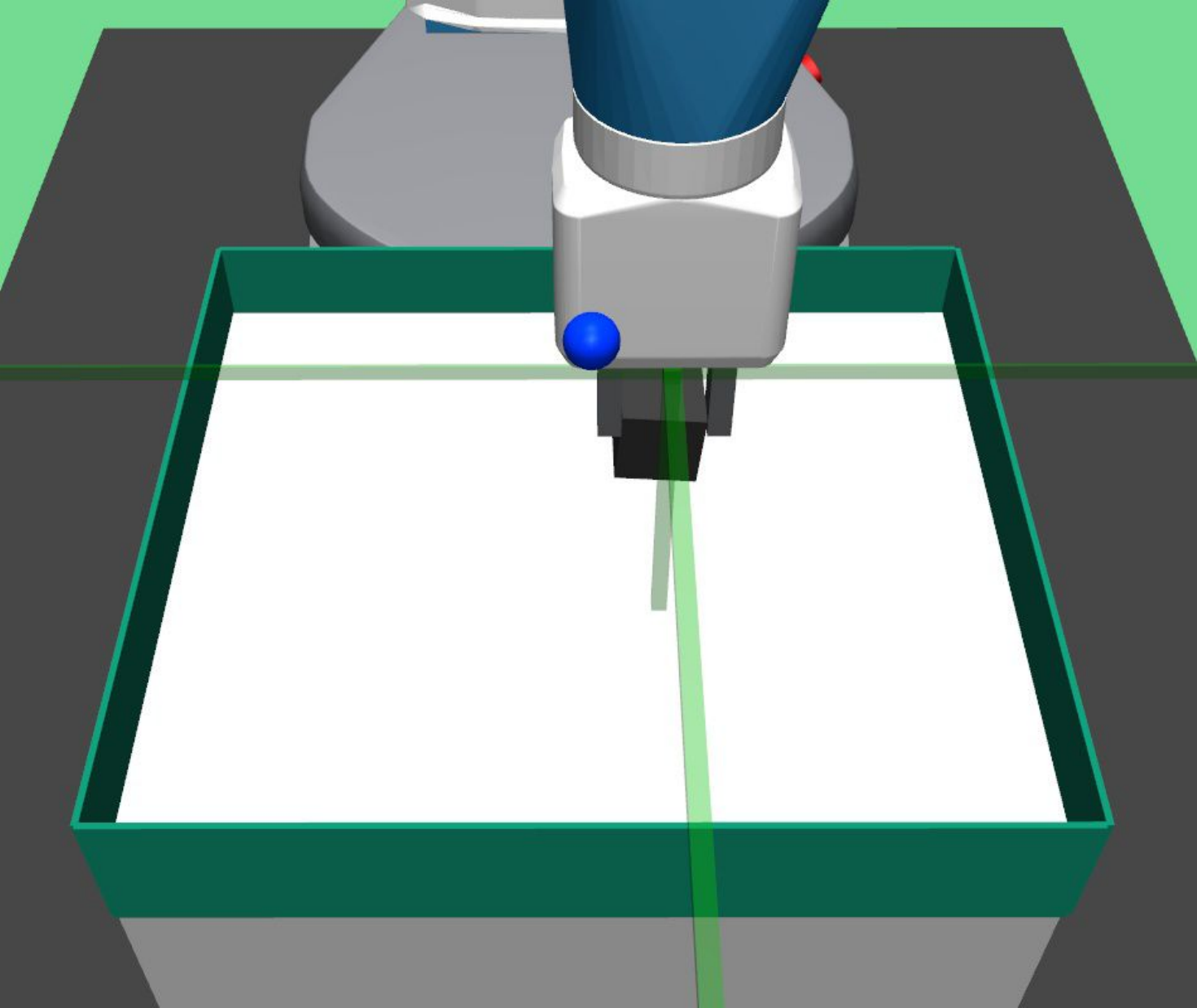}
\includegraphics[scale=0.09]{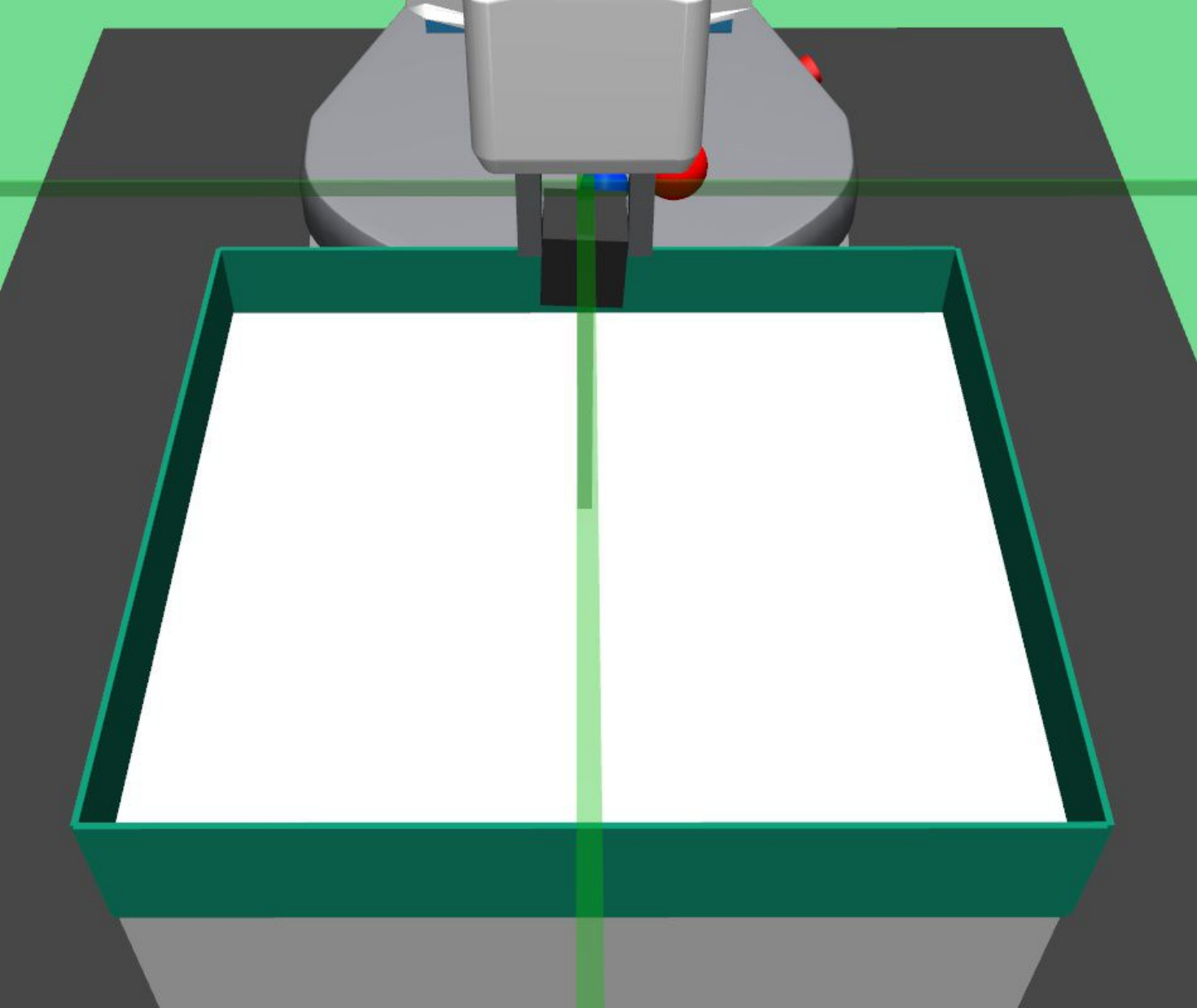}
\includegraphics[scale=0.09]{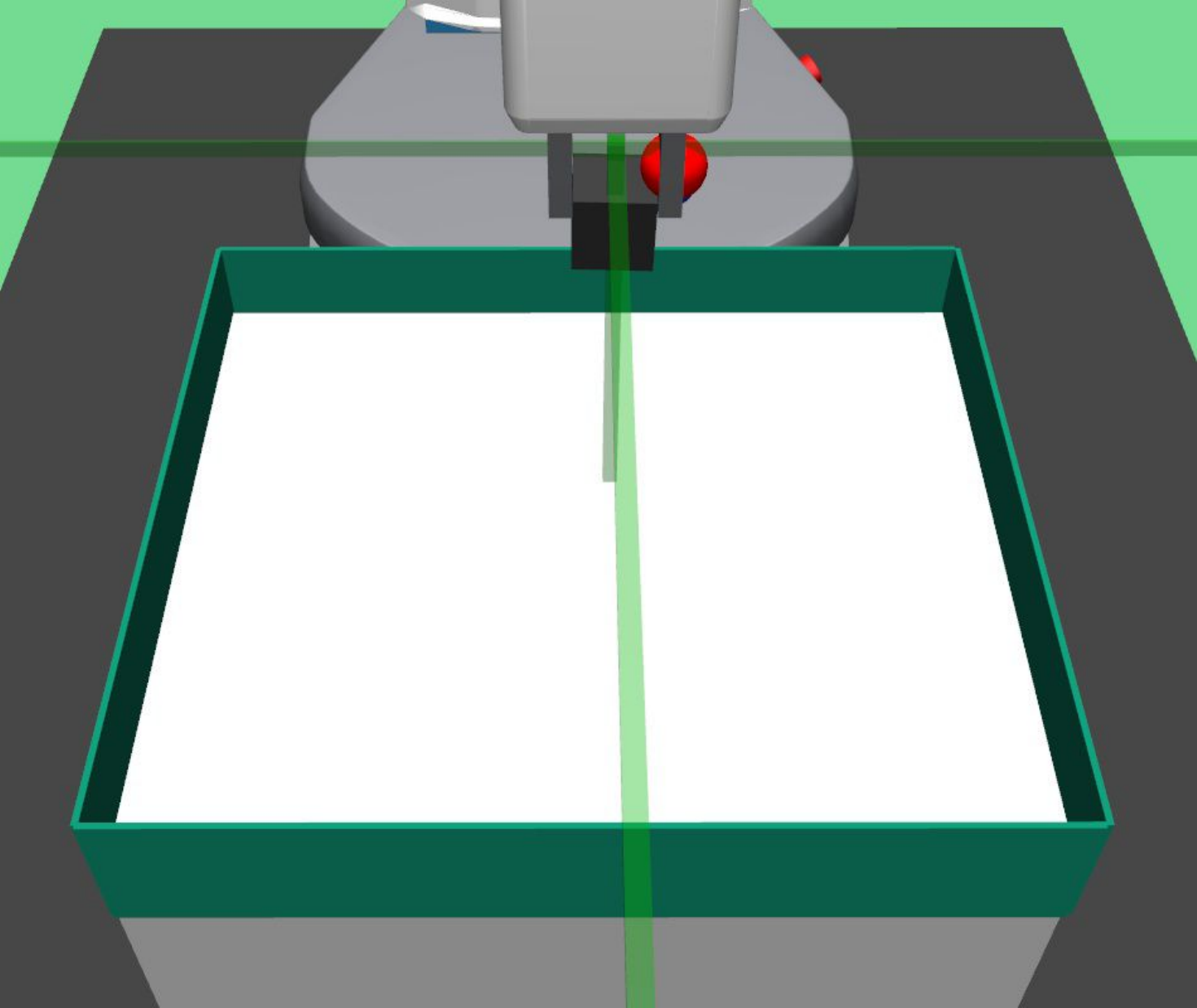}
\includegraphics[scale=0.09]{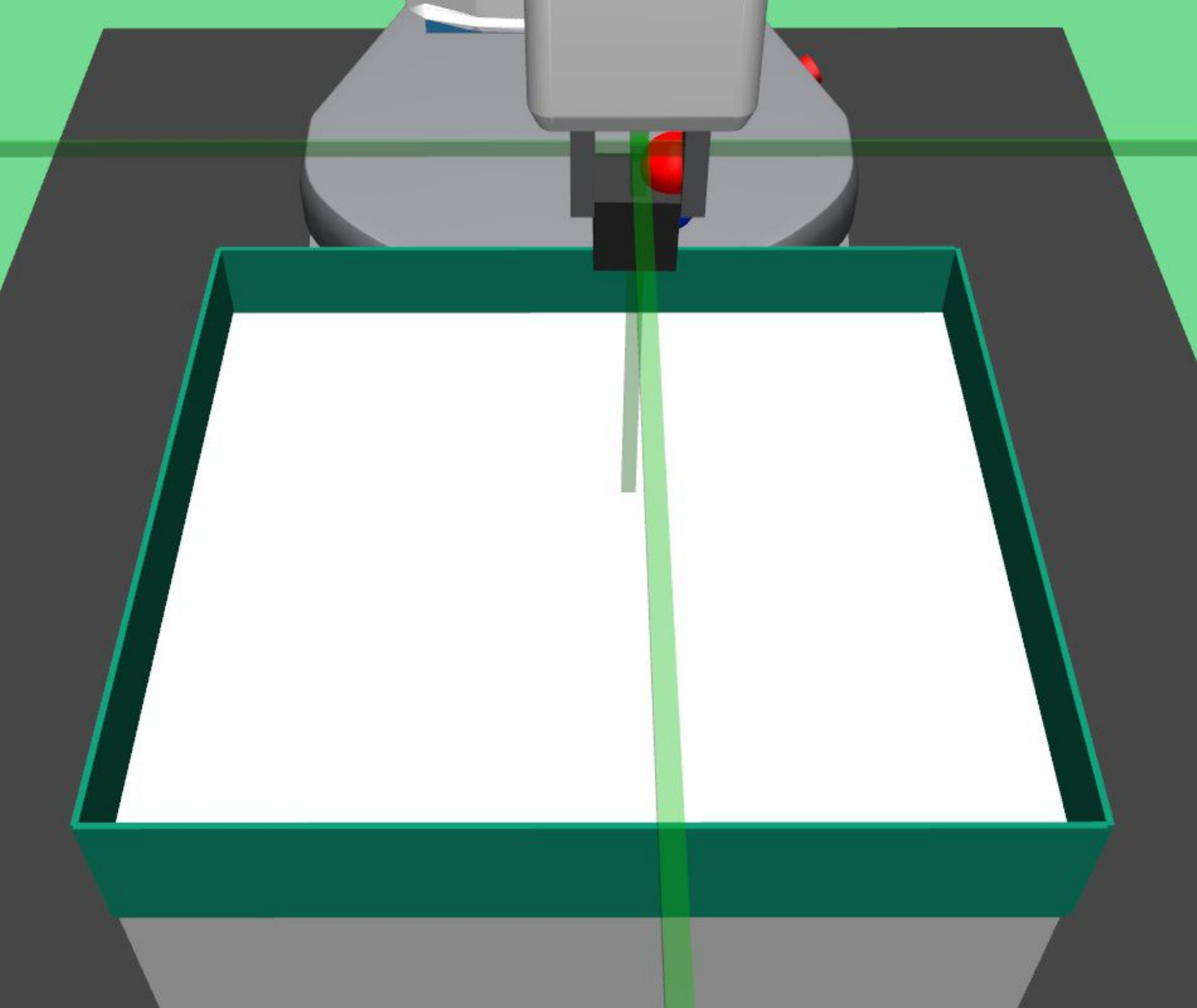}
\includegraphics[scale=0.09]{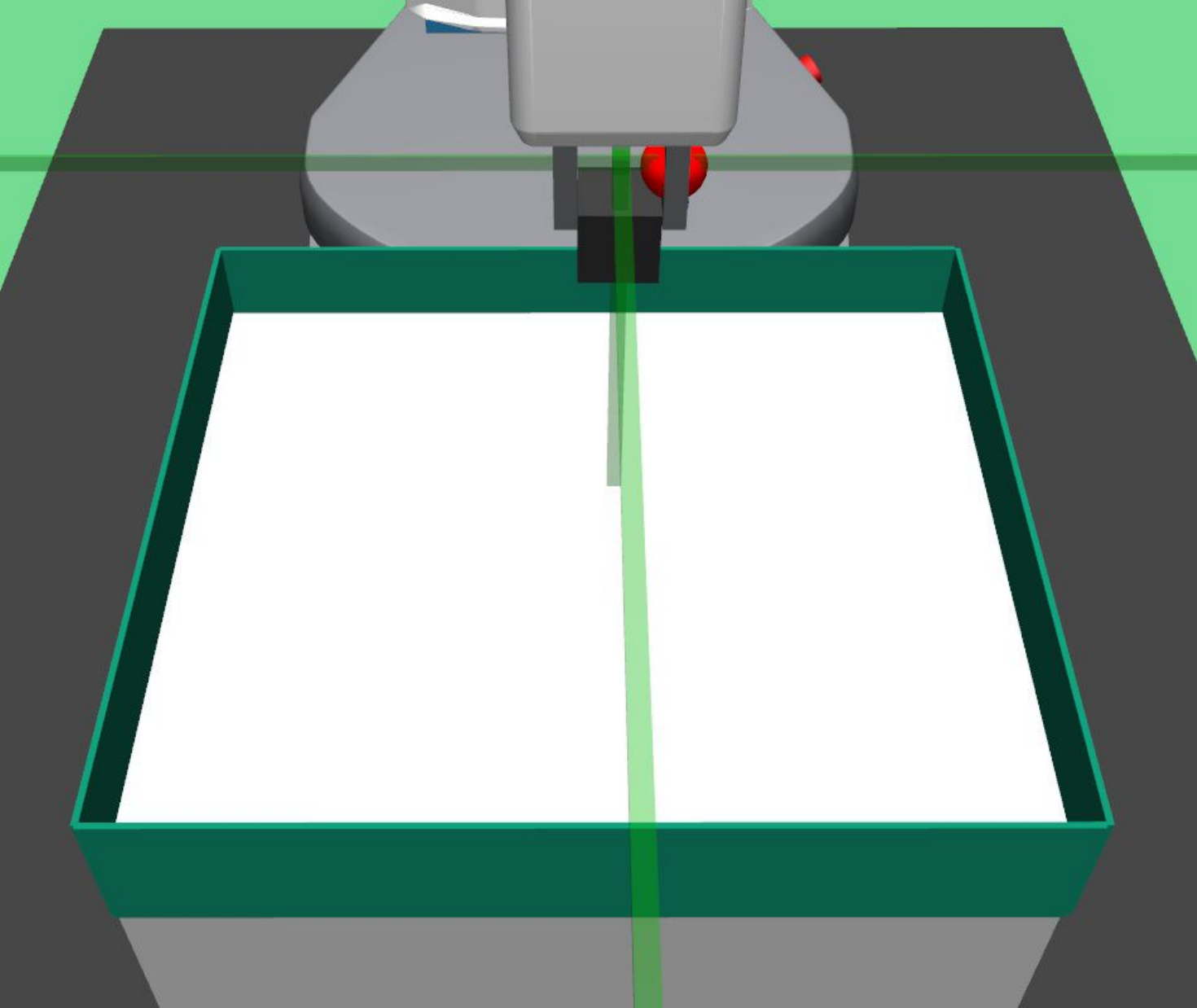}
\caption{\textbf{Successful visualization}: The visualization is a successful attempt at performing pick and place task.}
\label{fig:pick_viz_success_2_ablation}
\end{figure}

\begin{figure}[H]
\vspace{5pt}
\centering
\includegraphics[scale=0.075]{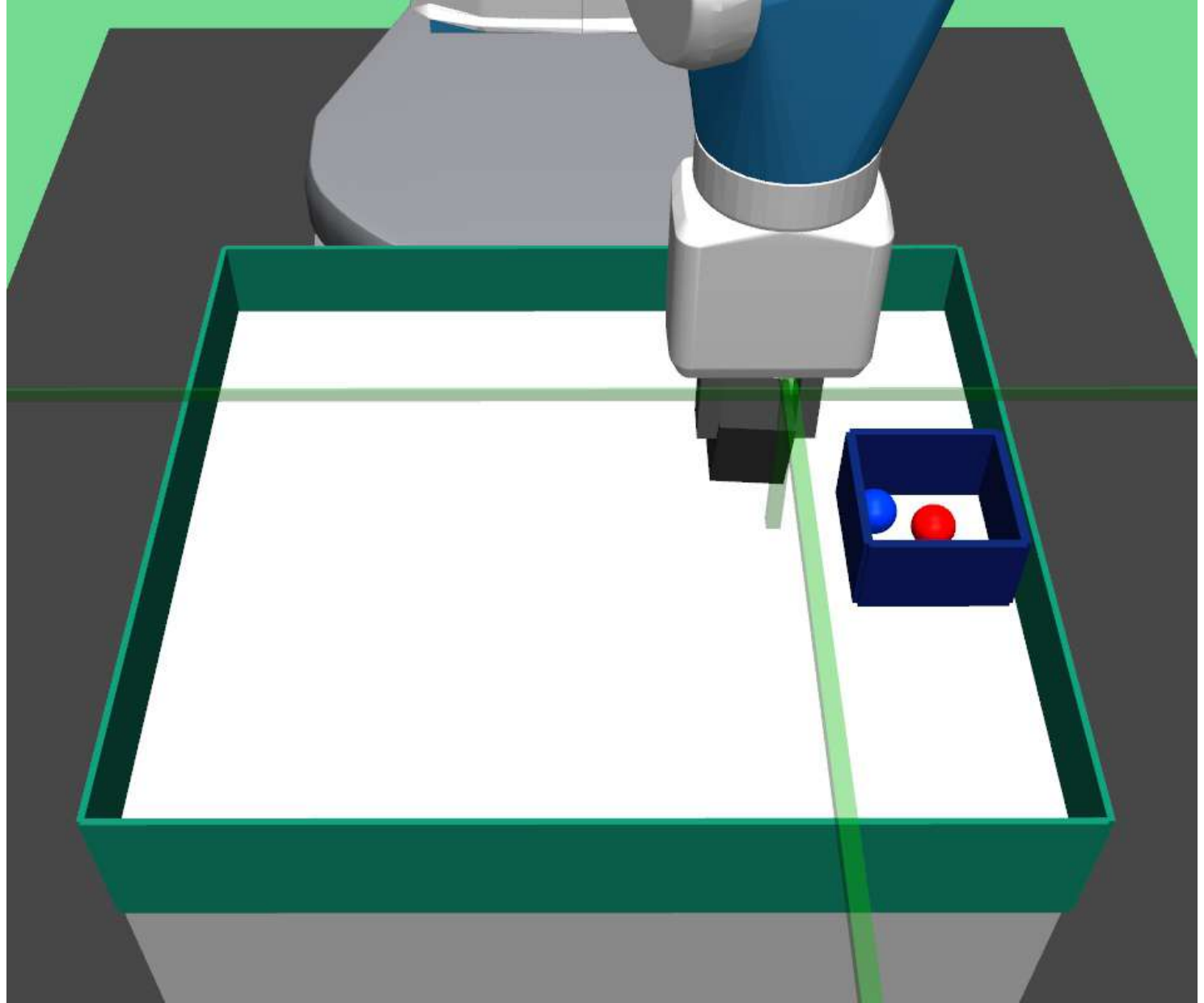}
\includegraphics[scale=0.075]{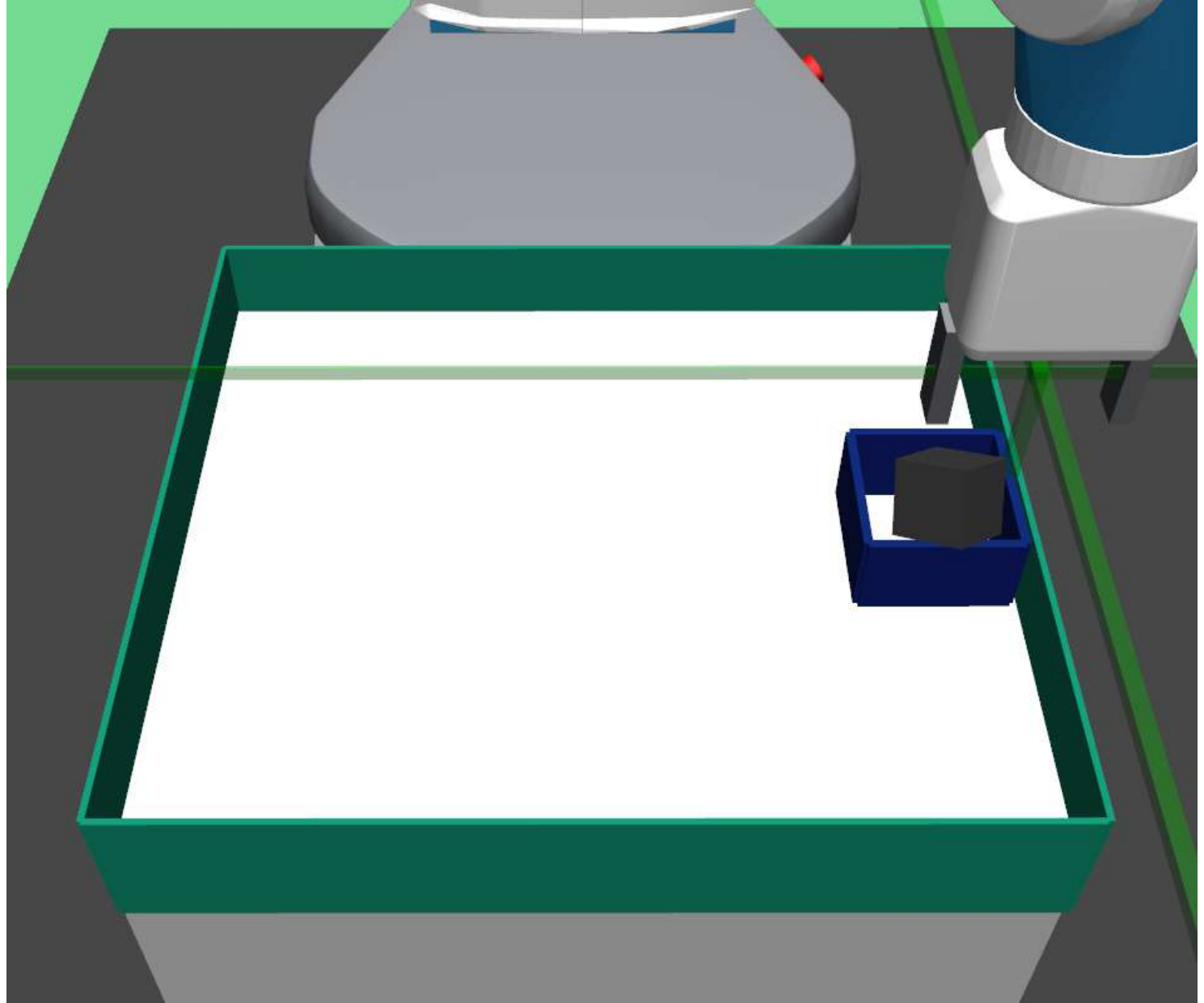}
\includegraphics[scale=0.075]{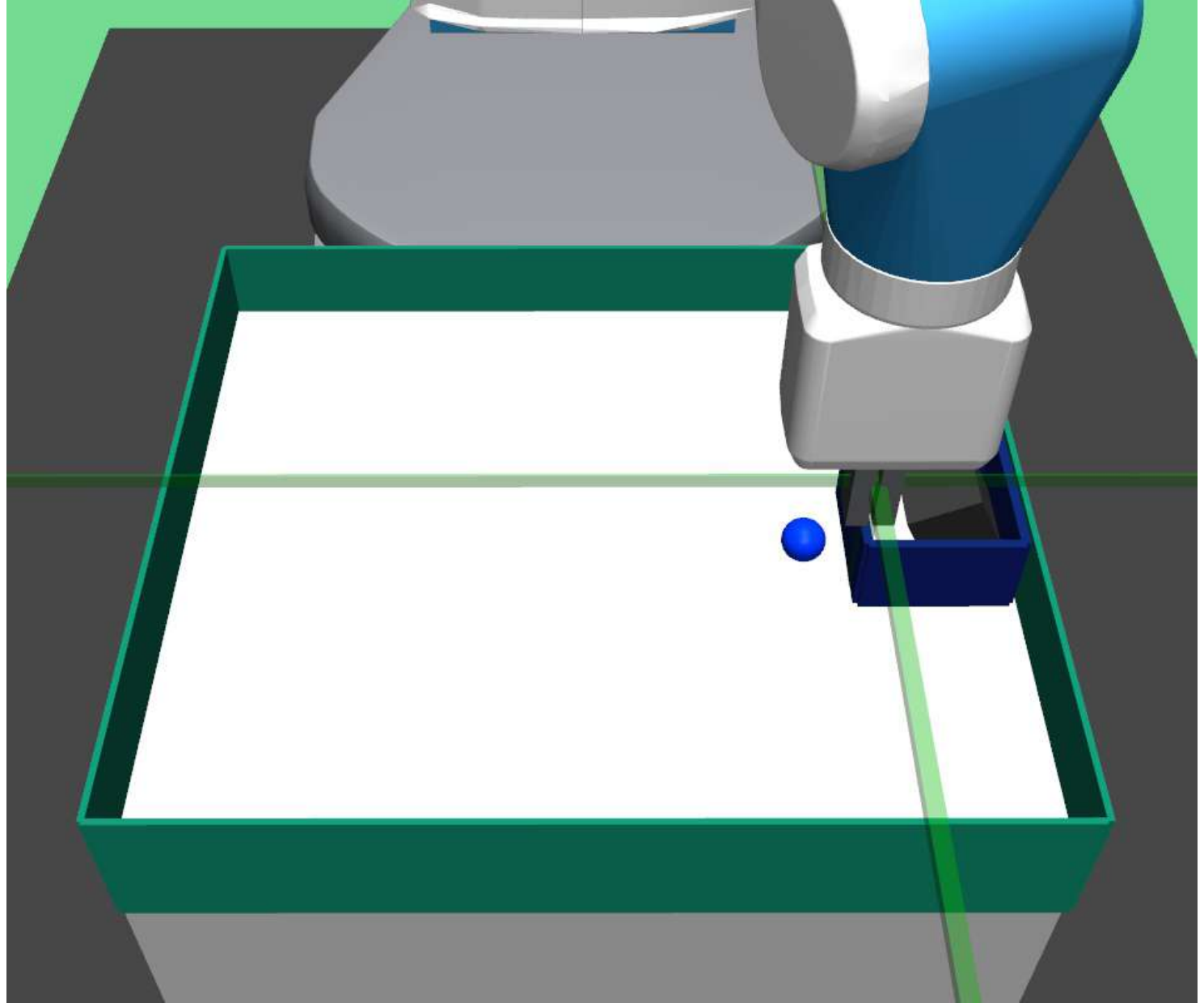}
\includegraphics[scale=0.075]{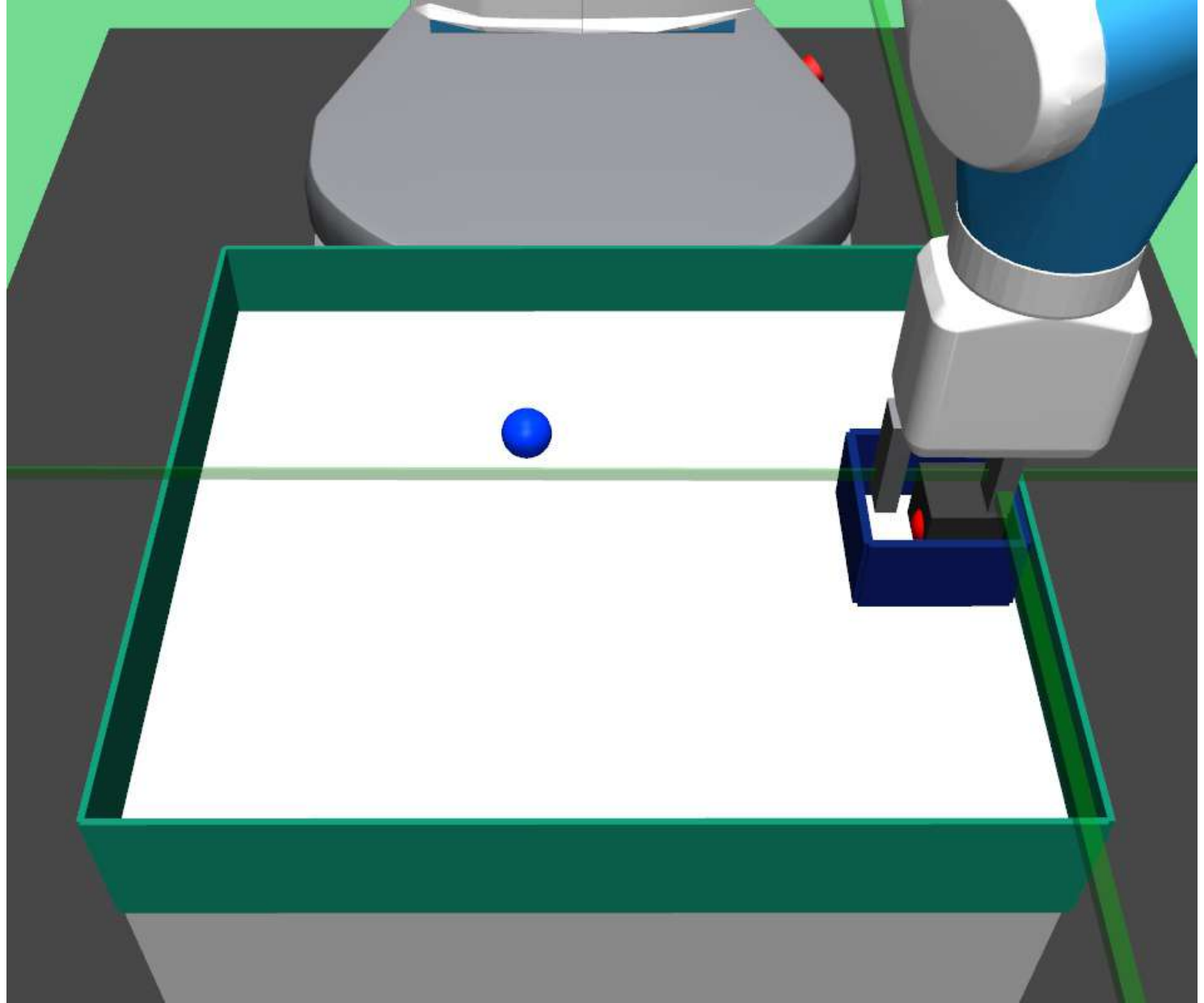}
\includegraphics[scale=0.075]{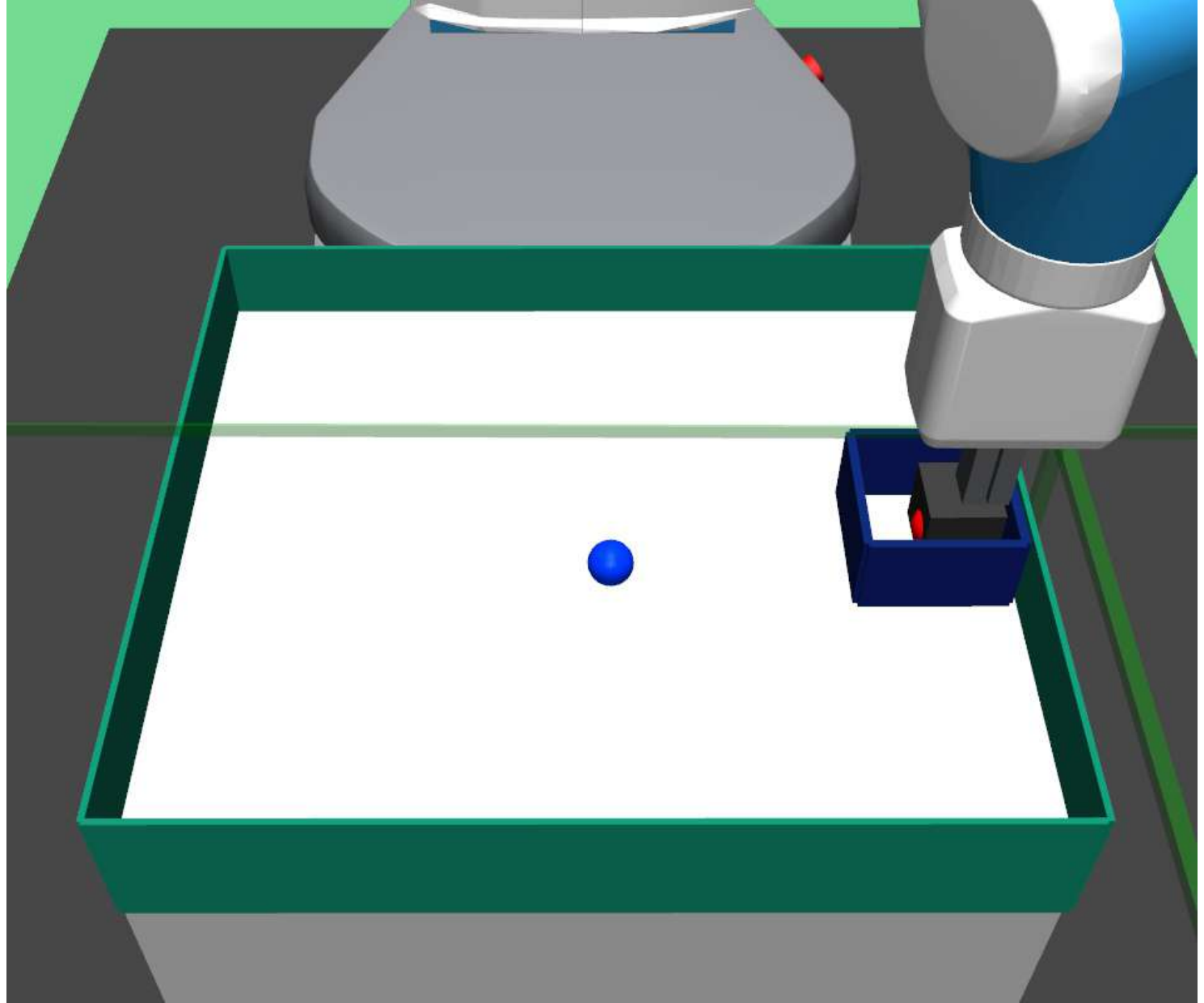}
\includegraphics[scale=0.075]{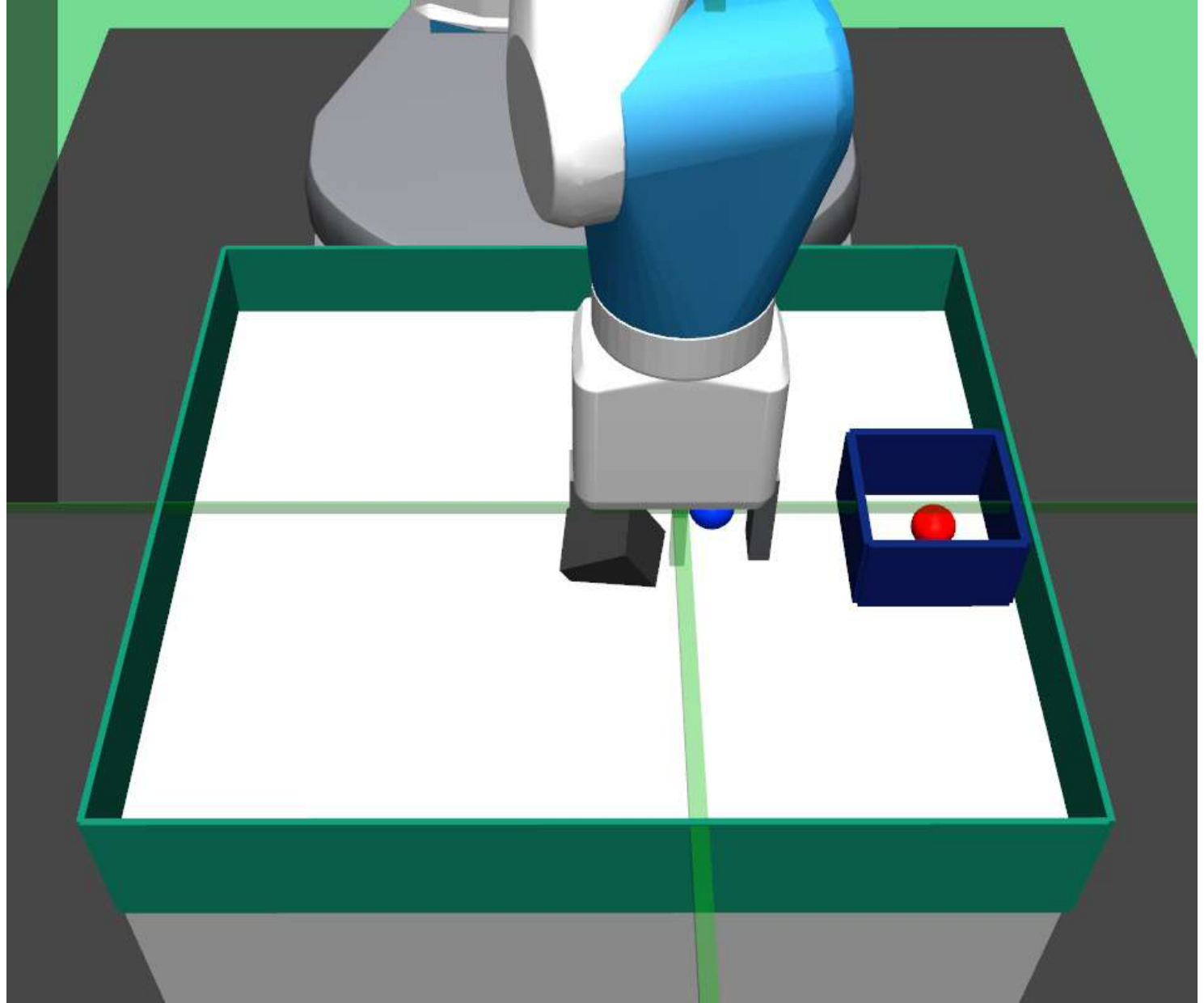}
\caption{\textbf{Successful visualization}: The visualization is a successful attempt at performing bin task.}
\label{fig:bin_viz_success_1_ablation}
\end{figure}

\begin{figure}[H]
\vspace{5pt}
\centering
\includegraphics[scale=0.08]{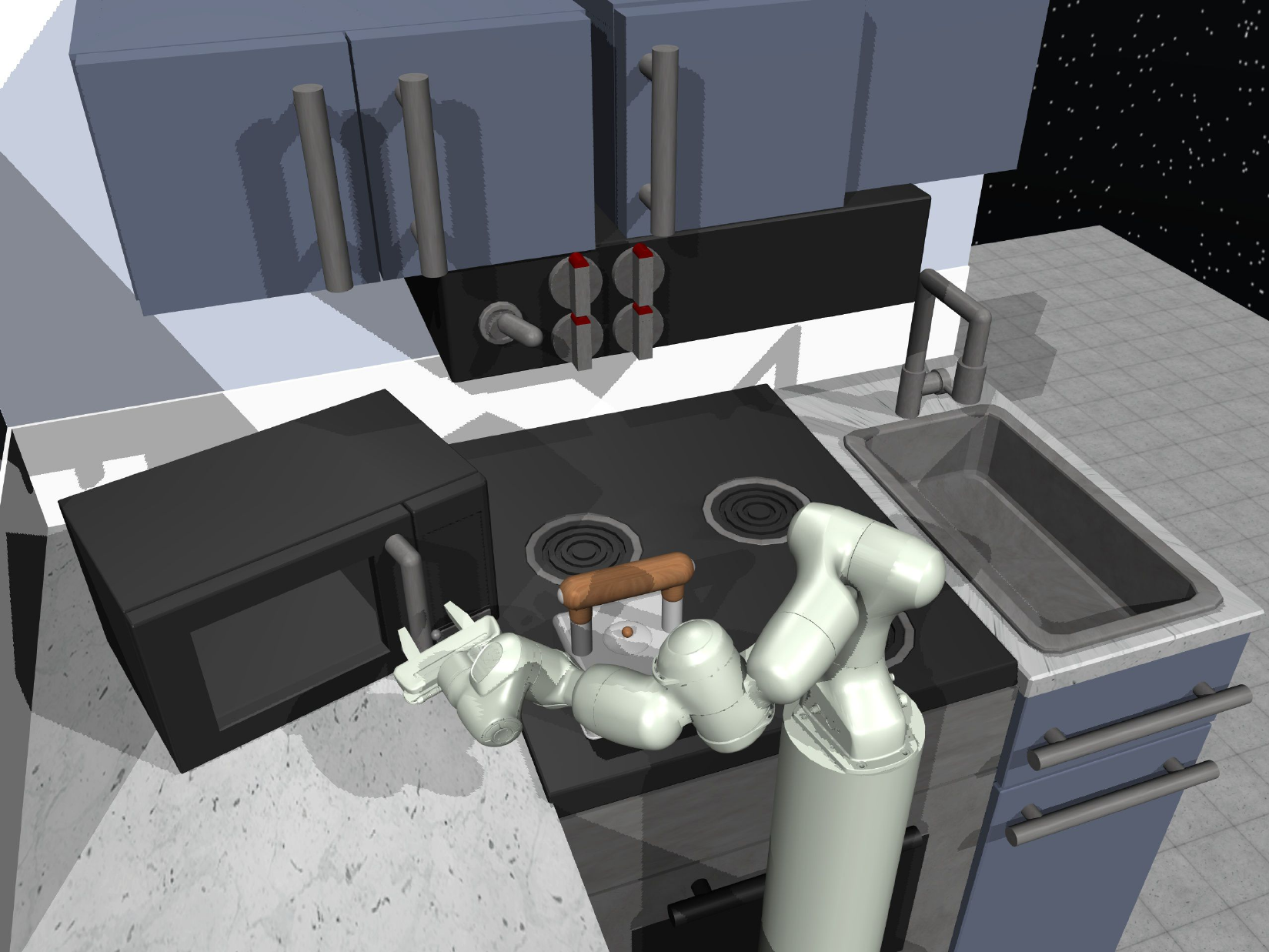}
\includegraphics[scale=0.08]{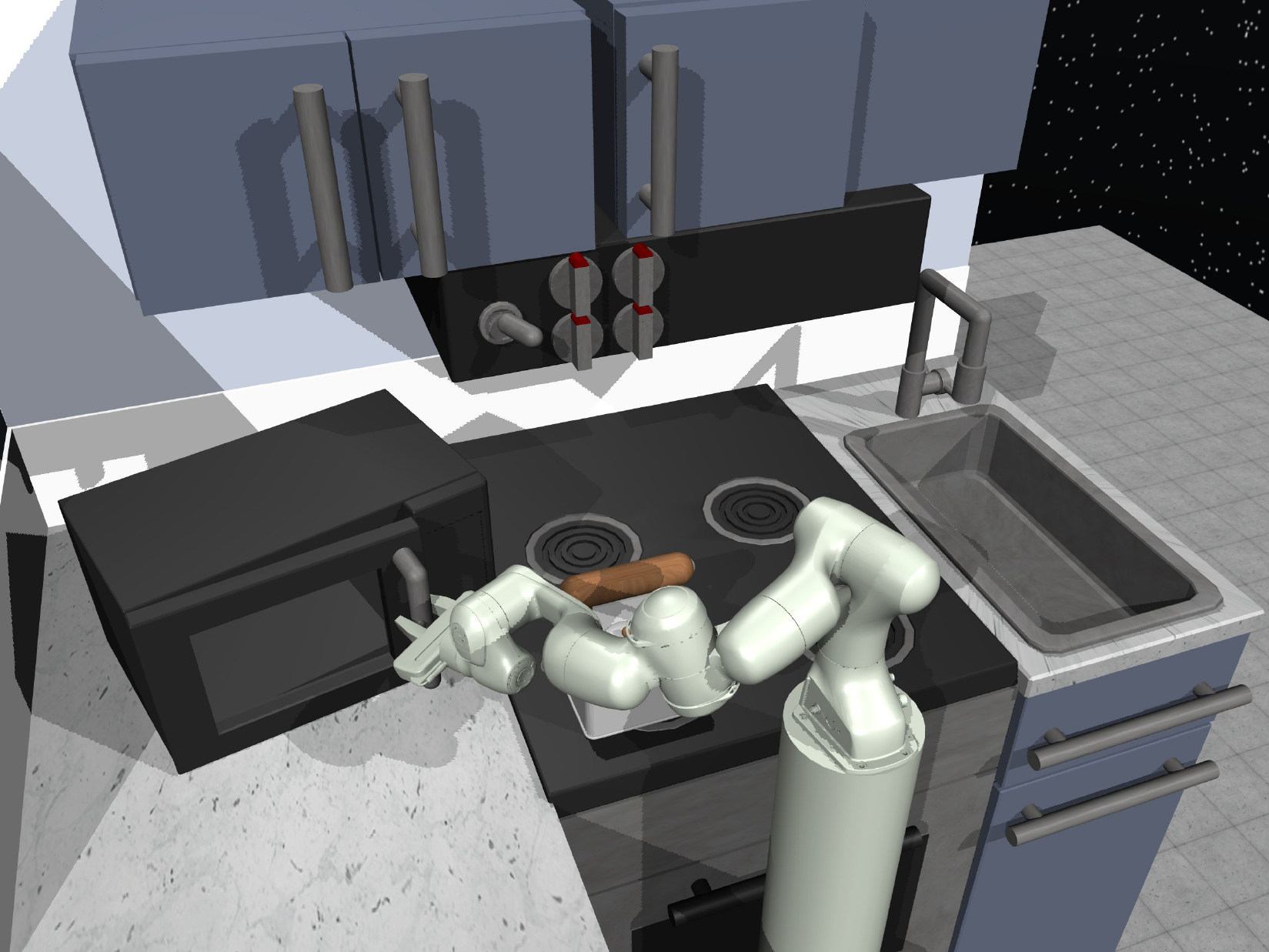}
\includegraphics[scale=0.08]{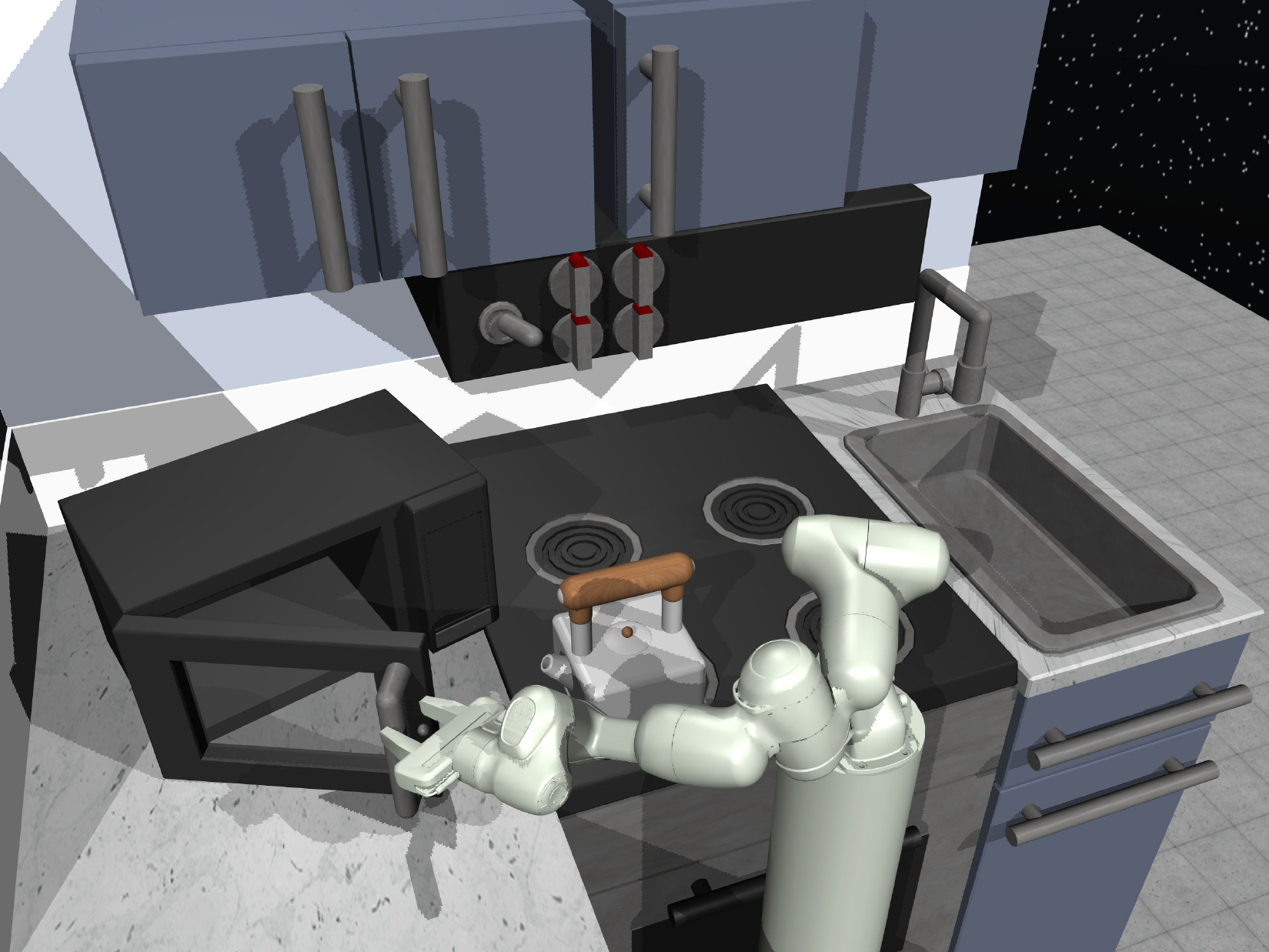}
\includegraphics[scale=0.08]{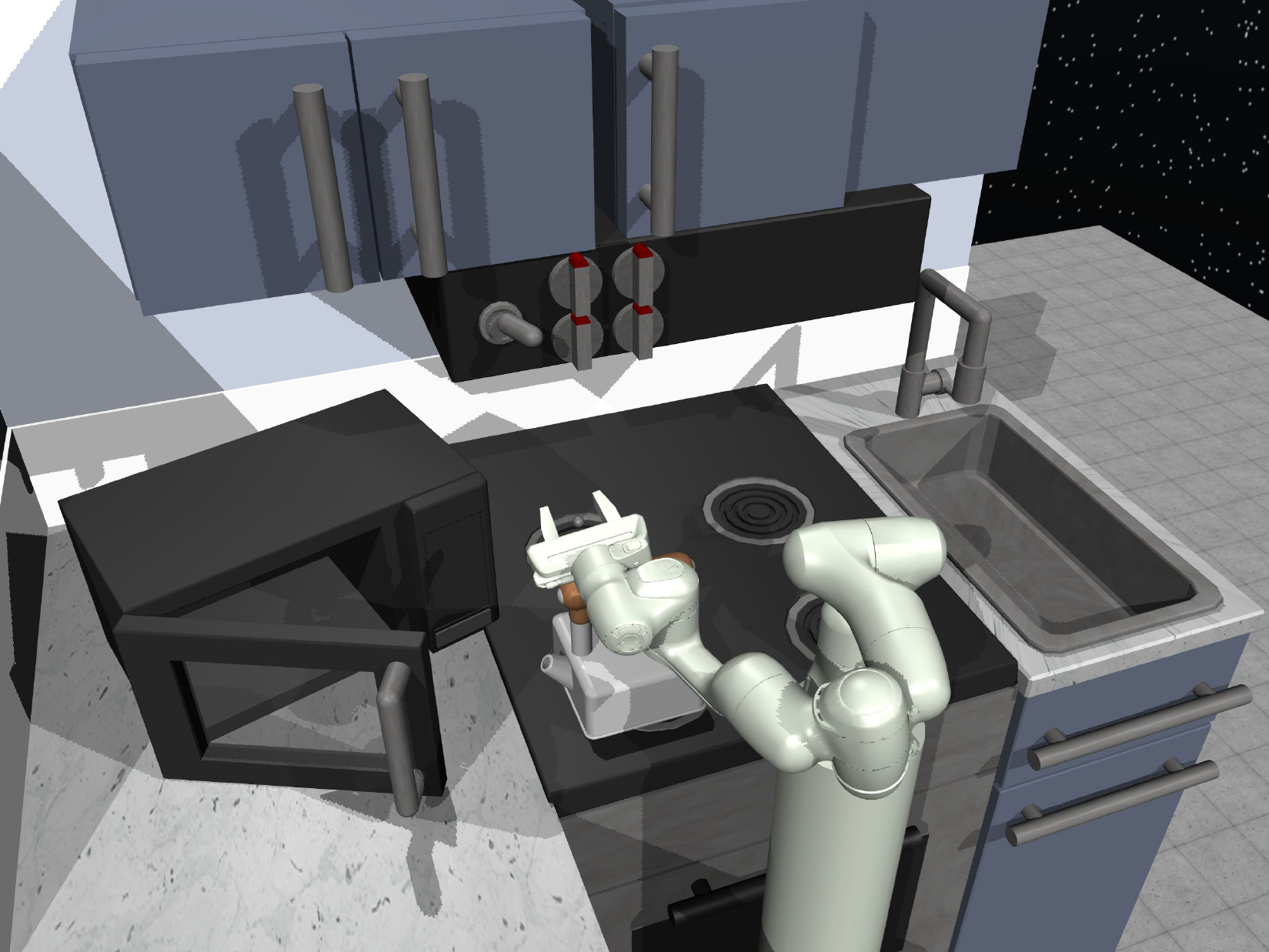}
\includegraphics[scale=0.08]{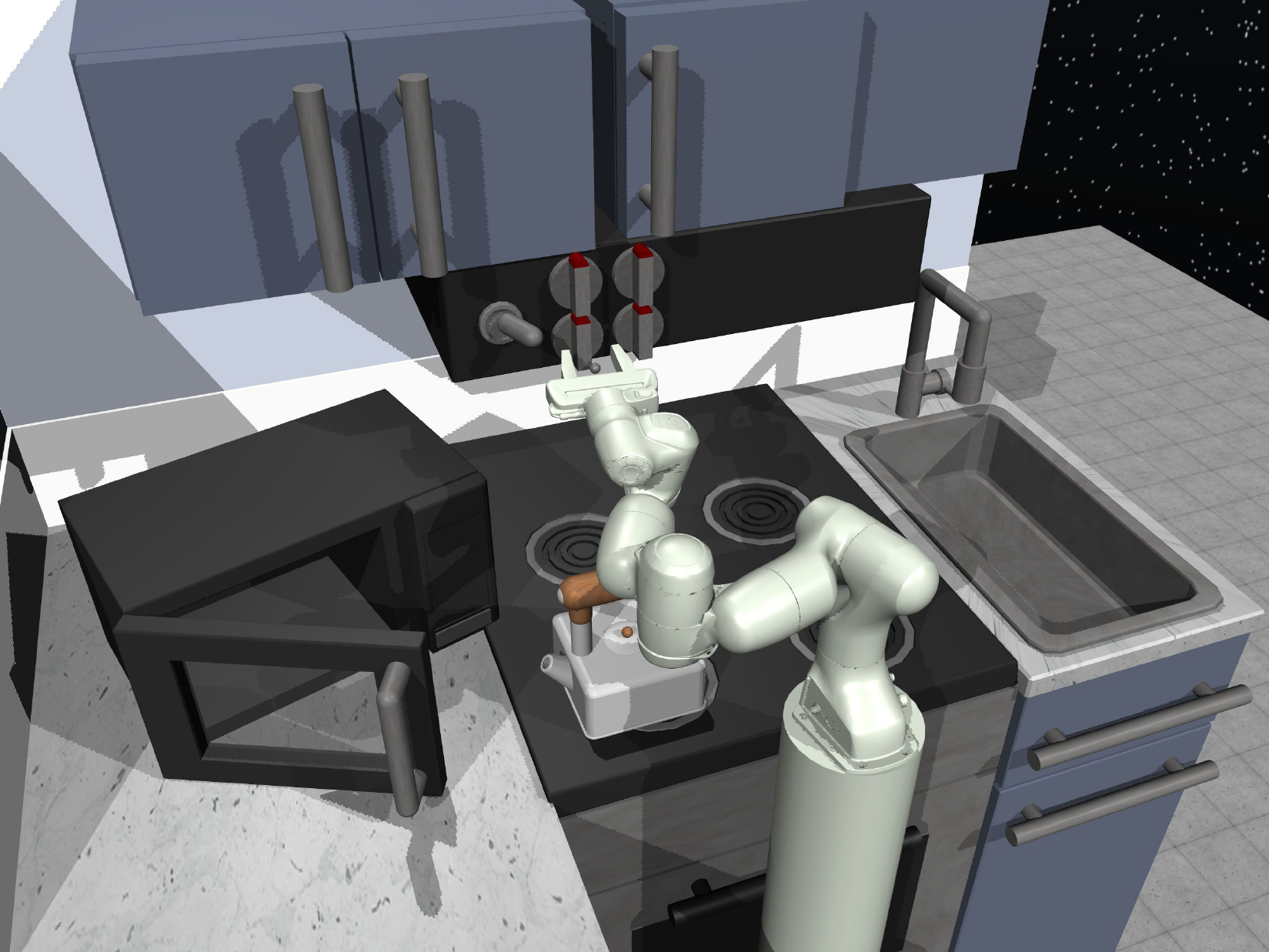}
\includegraphics[scale=0.08]{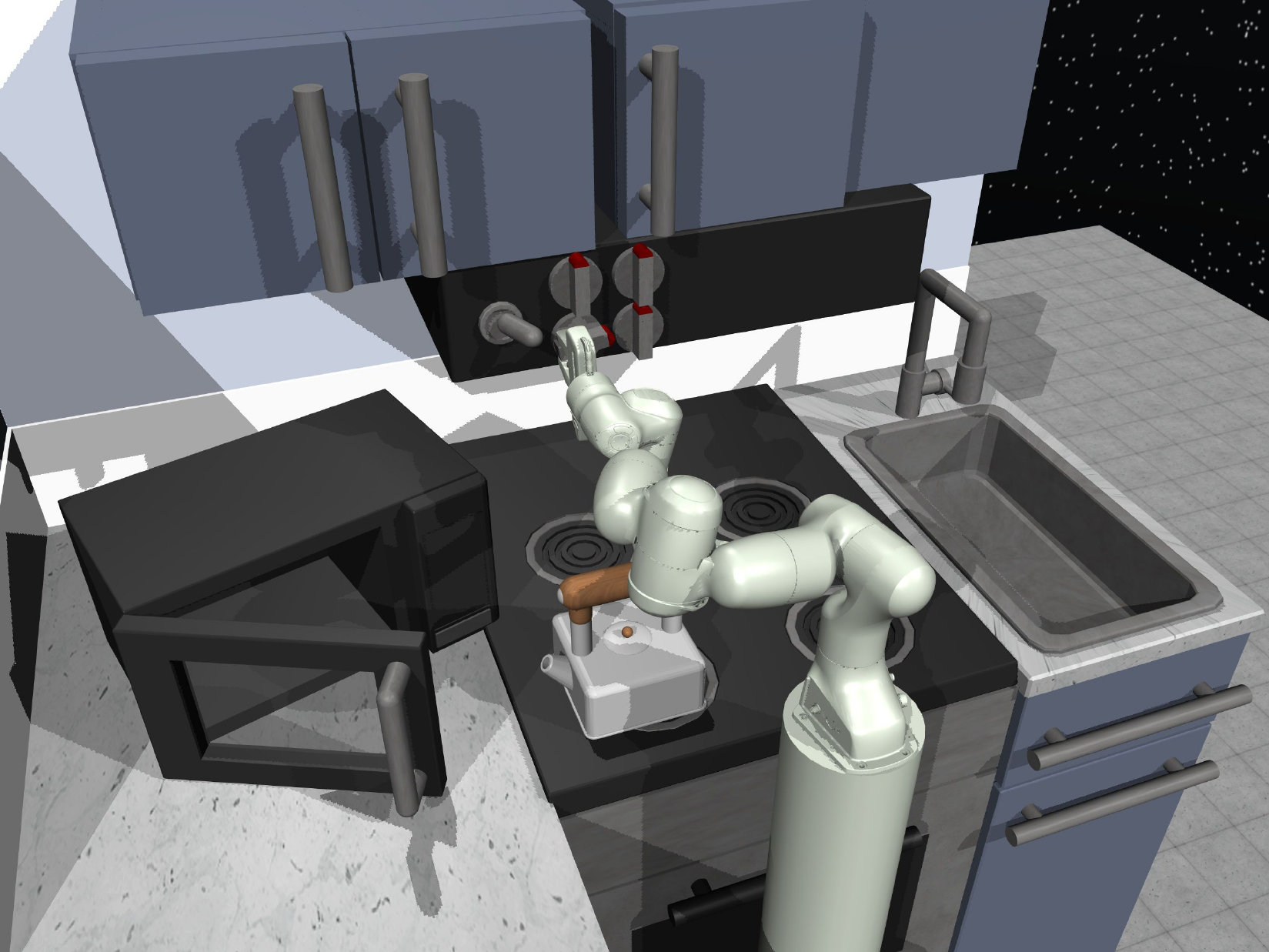}
\caption{\textbf{Successful visualization}: The visualization is a successful attempt at performing kitchen task.}
\label{fig:kitchen_viz_success_2_ablation}
\end{figure}

\end{document}